\documentclass{article}

\usepackage{arxiv}

\usepackage[utf8]{inputenc} 
\usepackage[T1]{fontenc}    

\usepackage{natbib} 
\usepackage{mathtools} 
\usepackage{amsfonts}
\usepackage{booktabs} 

\usepackage{algorithmic}
\usepackage{graphicx}
\usepackage{textcomp}
\usepackage{hyperref}

\usepackage{caption}
\usepackage{subcaption}
\usepackage{float}
\usepackage{bm} 
\usepackage{listings}
\usepackage[table]{xcolor}
\usepackage{geometry}
\usepackage{adjustbox} 
\usepackage{makecell}  
\usepackage{array}
\usepackage{authblk}
\usepackage{float}

\hypersetup{
    colorlinks=true,
    citecolor=blue,
    linkcolor=blue,
    urlcolor=blue
}

\title{Accelerating Simulation-Based Inference with Variational Autoencoders}

\author[1]{Mayank Nautiyal}
\author[2]{Andrey Shternshis}
\author[2]{Andreas Hellander}
\author[1]{Prashant Singh}

\affil[1]{Science for Life Laboratory, Uppsala University}

\affil[2]{Uppsala University}
\date{}
  
\begin{document}
\maketitle

\begin{abstract}
We present a generative modeling approach based on the variational inference framework for likelihood-free simulation-based inference. The method leverages latent variables within variational autoencoders to efficiently estimate complex posterior distributions arising from stochastic simulations. We explore two variations of this approach distinguished by their treatment of the prior distribution. The first model adapts the prior based on observed data using a multivariate prior network, enhancing generalization across various posterior queries. In contrast, the second model utilizes a standard Gaussian prior, offering simplicity while still effectively capturing complex posterior distributions. We demonstrate the ability of the proposed approach to approximate complex posteriors while maintaining computational efficiency on well-established benchmark problems.
\end{abstract}

The simulation-based inference (SBI) problem involves statistical inference of parameters \(\boldsymbol{\theta}\) of a stochastic simulation model \( M(\boldsymbol{\theta}, \xi) \) containing random states \(\xi\), from observed data \(\boldsymbol{y}\) \citep{Cranmer20}. 
Once calibrated, the simulation model can be used to reason about, analyze and explain the observed data \(\boldsymbol{y}\) in the context of the corresponding physical process . Several challenges arise in SBI owing to model stochasticity and potential multi-valuedness, where different sets of parameter values can produce similar observations, or similar parameters may lead to varied outputs. Consequently, the task is to infer the posterior distribution \( p(\boldsymbol{\theta} \mid \mathbf{y} ) \) in a Bayesian setting.


A user-defined prior distribution \( p(\boldsymbol{\theta}) \) encodes  domain knowledge or assumptions about plausible parameter values. 
The likelihood function \( p(\mathbf{y} \mid \boldsymbol{\theta}) \) is however, often intractable due to intricate stochastic dynamics, high-dimensional integration, or latent variables making analytical computation impractical \citep{sisson2018handbook}. Although a closed-form expression for the likelihood function \( p(\mathbf{y} \mid \boldsymbol{\theta}) \) is unavailable, samples from this distribution can be generated by evaluating the simulator \( M \) with varying \( \xi \) for different \( \boldsymbol{\theta} \) drawn from the prior, effectively yielding samples from the joint distribution \( (\boldsymbol{\theta}, \mathbf{y}) \sim p(\boldsymbol{\theta}, \mathbf{y}) \).


Deep learning based approaches that directly approximate the model likelihood, or the posterior distribution have gained increasing popularity in the SBI setting \citep{zammit2024neural}. Several approaches involve flow-based models, which are potentially compute-intensive and also restrict the design space of neural networks to invertible functions with computationally efficient Jacobian calculations \citep{flow}. This constraint can limit the flexibility of the model in handling more complex data structures. Alternative approaches like Generative Adversarial Training for SBI (GATSBI) \citep{Ramesh22} use an adversarial network, where a generator produces simulations that mimic observed data, and a discriminator distinguishes between real and simulated data. 
However, the adversarial approach can suffer from training instability, mode collapse, and difficulties in balancing the generator and discriminator networks \citep{Arjovsky17a, Arjovsky17b}, complicating training and limiting robustness. Recent approaches such as Simformer \citep{gloecklerall} combine diffusion models with transformers and cross attention to enable highly accurate posterior estimation, but entail relatively high computational cost for training/inference.

{\bf Contribution: }We propose a variational inference approach for simulation-based inference, utilizing a Conditional Variational Autoencoder (C-VAE) architecture \citep{kingma2022autoencodingvariationalbayes, NIPS2015_8d55a249} that is simple, computationally efficient, interpretable and robust. As VAEs are not constrained by the invertibility requirement, they allow seamless integration of various neural network architectures without significant modifications as opposed to flow-based models. VAEs also avoid the instability associated with GANs due to adversarial learning dynamics. As a result, C-VAEs can handle complex data structures and dependencies more effectively, making them suitable for a wider range of tasks without requiring external summary networks \citep{Cranmer20}. The simplicity and efficiency also makes the approach particularly suitable for heavy workflows such as model exploration, which entail repeated solution of several inference problems. 

The paper is organized as follows. Section \ref{sec:related} outlines the related work. In Section \ref{Problem Formulation}, we provide a formal problem definition, followed by a detailed explanation of the proposed architecture in Sections \ref{model_a} and \ref{model_b}. A comprehensive comparison on benchmark test problems is presented in Section \ref{Experiments}. Section \ref{sec:discussion} discusses the results, while Section \ref{Conclusion} concludes the paper by summarizing the key findings and further advancements.

\section{Related Work}
\label{sec:related}
Recent deep learning advances have spurred a variety of SBI approaches. There exist methods that focus on approximating the likelihood directly \citep{hermans2020likelihoodfree}, while others concentrate on learning the posterior function \citep{papamakarios2016fast, NIPS2017_addfa9b7, Greenberg19}. A growing line of research pursues hybrid strategies that estimate both likelihood and posterior \citep{Radev23,glöckler2022variational,wiqvist2021sequential}, aiming to balance flexibility and computational efficiency.

Neural Posterior Estimation (NPE) \citep{papamakarios2016fast} was one of the earliest approaches towards one-step posterior estimation, introducing the use of conditional normalizing flows to approximate the posterior distribution. By incorporating techniques such as Neural Spline Flows (NSFs) and Mixture Density Networks (MDNs), NPE enhanced both the accuracy and scalability of simulation-based inference. In comparison to traditional approaches like Approximate Bayesian Computation (ABC) or Sequential Monte-Carlo (SMC)-ABC \citep{sisson2018handbook} that rely on repeated simulations and rejection-based schemes, NPE and its derivatives such as Robust NPE \citep{Ward22} are more efficient and flexible.

Sequential Neural Posterior Estimation (SNPE) improves NPE's efficiency by employing sequential training, where initial parameters are sampled from the prior, and the model is iteratively refined by focusing on regions of the parameter space that are most relevant to the parameter inference task. This adaptive sampling allows for more efficient exploration of the posterior distribution. Early versions like SNPE-A \citep{papamakarios2016fast} and SNPE-B \citep{NIPS2017_addfa9b7} required correction steps to account for changes in the sampling distribution, adding complexity to the process. In contrast, Automatic Posterior Transformation (APT) or SNPE-C \citep{Greenberg19}, eliminated this need, offering a more robust and streamlined approach by leveraging normalizing flows to refine the posterior iteratively. However, a limitation of sequential methods like APT is that they entail multiple rounds of training, extending the total training time. 

Jointly Amortized Neural Approximation (JANA) framework \citep{Radev23} addresses posterior and likelihood estimation through joint amortization. JANA employs conditional invertible neural networks (cINNs) for both posterior and likelihood networks, allowing efficient transformations between the parameter space and latent variables. Additionally, JANA includes a trainable summary network sub-module optimized to extract maximally informative data representations in an end-to-end manner. Unlike previous approaches, JANA follows a fully amortized strategy, allowing the evaluation of normalized densities and generation of conditional random draws for parameter estimation and surrogate modeling. 

The recently proposed Simformer \citep{gloecklerall} employs a score-based diffusion model combined with transformer architectures and a cross-attention conditioning mechanism, allowing it to accurately learn complex, high-dimensional posteriors with substantial flexibility. However, diffusion-based approaches generally require iterative reverse-diffusion sampling, which, combined with the transformer’s computational overhead, results in slower amortized inference for higher-dimensional problems compared to other generative models.

We compare our proposed approach with five popular SBI methods: JANA \citep{Radev23}, GATSBI \citep{Ramesh22}, APT \citep{Greenberg19},NPE \citep{papamakarios2016fast} and Simformer \citep{gloecklerall}. The motivation is to cover a variety of approaches (e.g., flow-based, adversarial, diffusion, one-shot, sequential) while noting that our focus is on efficiency and simplicity as opposed to solely the posterior approximation accuracy. We refer the reader to \citep{zammit2024neural} and \citep{Radev23} for further reading on deep learning methods for SBI.

\section{Simulation Based Inference}
\label{Problem Formulation}
\noindent
In the Bayesian framework, the aim is to infer the posterior distribution \( p(\boldsymbol{\theta} \mid \mathbf{y}) \) over the model parameters \( \boldsymbol{\theta} \), given the observed data  \(\mathbf{y}\). From Bayes’ theorem, the posterior distribution can be expressed as:
\begin{equation}
p(\boldsymbol{\theta} \mid \mathbf{y}) = \frac{p(\mathbf{y} \mid \boldsymbol{\theta}) p(\boldsymbol{\theta})}{p(\mathbf{y})},
\label{eq:posterior}
\end{equation}
where \( p(\mathbf{y}) = \int p(\mathbf{y} \mid \boldsymbol{\theta}) \, p(\boldsymbol{\theta}) \, d\boldsymbol{\theta} \) is the marginal likelihood, which is a normalizing constant ensuring $p(\boldsymbol{\theta} \mid \mathbf{y})$ is a valid probability distribution. However, as \( p(\mathbf{y}) \) is independent of \( \boldsymbol{\theta} \), it can be ignored in the computations involving the posterior. 

We propose two models for amortized inference: a Conditional Prior VAE (CP-VAE) and an Unconditional Prior VAE (UP-VAE). Both models utilize latent variables to capture hidden structures in the data. The Conditional Prior VAE uses a multivariate prior network that conditions on observed data, enabling adaptive inference that improves generalization across posterior queries. The Unconditional Prior VAE, in contrast, employs a simpler Gaussian prior with unit variance, while still capturing complex posterior distributions. Leveraging the variational inference framework, both models achieve competitive performance across a range of tasks, including bimodal and high-dimensional settings.

\subsection{Conditional Prior VAE (CP-VAE)}
\label{model_a}
The Conditional Prior Variational Autoencoder (CP-VAE) extends the standard Conditional Variational Autoencoder (CVAE) \citep{NIPS2015_8d55a249,ivanov2018variational} specifically designed for simulation-based inference. We begin by introducing an auxiliary latent variable \( \mathbf{z} \) to capture complex structures and dependencies within the joint distribution \(p(\boldsymbol{\theta}, \mathbf{y})\) of parameters \(\boldsymbol{\theta}\) and simulated data \(\mathbf{y}\). The latent variable \(\mathbf{z}\) serves as a learned, low-dimensional summary statistic, efficiently encoding the most salient features of high-dimensional data relevant to characterize the posterior distribution. We formulate a latent variable model \( p(\mathbf{z} \mid \mathbf{y}, \boldsymbol{\theta}) \) and approximate it using a variational distribution \( q_{\psi}(\mathbf{z} \mid \mathbf{y}, \boldsymbol{\theta}) \), parameterized by \( \psi \in \Phi \), where \( \Phi \) represents a family of distributions. The goal is to minimize the Kullback-Leibler (KL) divergence between this variational distribution and the true conditional distribution \citep{jordan1999introduction, kingma2022autoencodingvariationalbayes}:
\begin{equation}
\min_{\psi \in \Phi} D_{KL}\bigg(q_{\psi}(\mathbf{z} \mid \mathbf{y}, \boldsymbol{\theta}) \| p(\mathbf{z} \mid \mathbf{y}, \boldsymbol{\theta})\bigg).
\label{eq:KL_latent}
\end{equation}
Using Bayes' theorem and the chain rule of probability, the conditional distribution \(p(\mathbf{z} \mid \mathbf{y}, \boldsymbol{\theta})\) becomes:
\begin{equation}
p(\mathbf{z} \mid \mathbf{y}, \boldsymbol{\theta}) = \frac{p(\boldsymbol{\theta} \mid \mathbf{z}, \mathbf{y}) p(\mathbf{z} \mid \mathbf{y})}{p(\boldsymbol{\theta} \mid \mathbf{y})}.
\label{eq:conditional_distribution}
\end{equation}
Substituting \eqref{eq:conditional_distribution} into \eqref{eq:KL_latent}, and denoting the KL divergence \(D_{KL}(q_{\psi}(\mathbf{z} \mid \mathbf{y}, \boldsymbol{\theta}) \| p(\mathbf{z} \mid \mathbf{y}, \boldsymbol{\theta}))\) as \(\mathcal{D}\), we obtain:
\begin{equation}
\begin{aligned}
    \mathcal{D}& = \mathbb{E}_{\mathbf{z} \sim q_{\psi}(\mathbf{z} \mid \mathbf{y}, \boldsymbol{\theta})} \bigg[ \log q_{\psi}(\mathbf{z} \mid \mathbf{y}, \boldsymbol{\theta}) - \log p(\mathbf{z} \mid \mathbf{y}) \\
    &\qquad - \log p(\boldsymbol{\theta} \mid \mathbf{y}, \mathbf{z}) + \log p(\boldsymbol{\theta} \mid \mathbf{y}) \bigg].
\end{aligned}
\label{eq:kl_divergence_latent}
\end{equation}
Rearranging the components of equation \eqref{eq:kl_divergence_latent}, we isolate \(\log p(\boldsymbol{\theta} \mid \mathbf{y})\) to derive:
\begin{equation}
\begin{aligned}
\log p(\boldsymbol{\theta} \mid \mathbf{y}) \geq & -D_{KL}\bigg(q_{\psi}(\mathbf{z} \mid \mathbf{y}, \boldsymbol{\theta}) \parallel p(\mathbf{z} \mid \mathbf{y})\bigg) + \mathbb{E}_{\mathbf{z} \sim q_{\psi}(\mathbf{z} \mid \mathbf{y}, \boldsymbol{\theta})}\bigg[\log p(\boldsymbol{\theta} \mid \mathbf{y}, \mathbf{z})\bigg].
\end{aligned}
\label{eq:elbo}
\end{equation}
This inequality stems from the fundamental non-negativity property of the KL divergence, establishing a lower bound on the log-posterior \( p(\boldsymbol{\theta} \mid \mathbf{y}) \). This bound forms the core of the optimization objective in the CP-VAE framework. Maximizing the log-posterior can be conveniently reformulated as minimizing its negative, which is a standard approach in optimization, leading to the following formulation:
\begin{equation}
\begin{aligned}
\mathcal{L}(\boldsymbol{\theta}, \mathbf{y}; \psi) = D_{KL}\bigg(q_{\psi}(\mathbf{z} \mid \mathbf{y}, \boldsymbol{\theta}) \,\|\, p(\mathbf{z} \mid \mathbf{y})\bigg)
- \mathbb{E}_{\mathbf{z} \sim q_{\psi}(\mathbf{z} \mid \mathbf{y}, \boldsymbol{\theta})} \bigg[\log p(\boldsymbol{\theta} \mid \mathbf{y}, \mathbf{z})\bigg].
\end{aligned}
\label{eq:optimization_problem_min}
\end{equation}

\begin{figure}[!t]
    \centering
    \includegraphics[width=6.5cm]{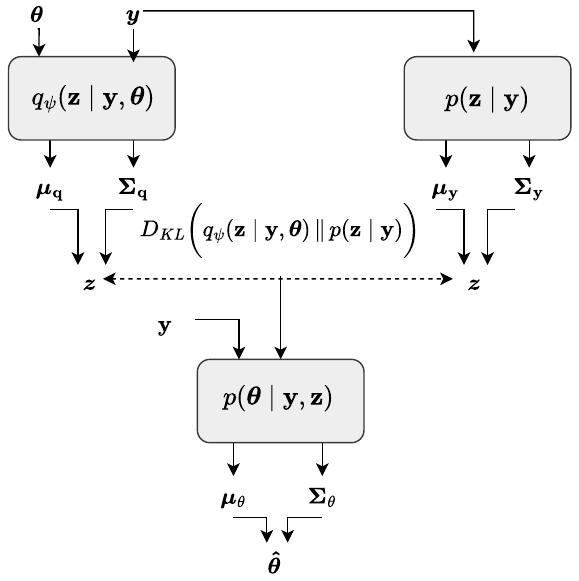}
    \caption{CP-VAE Architecture. The encoder \( q_{\psi}(\mathbf{z} \mid \mathbf{y}, \boldsymbol{\theta}) \) computes the parameters of the variational distribution for latent variables \(\mathbf{z}\), conditioned on paired simulated data \(\mathbf{y}\) and parameters \(\boldsymbol{\theta}\). The prior network \( p(\mathbf{z} \mid \mathbf{y}) \) defines a data-dependent prior over \(\mathbf{z}\), enforcing latent space regularization via the KL divergence. The decoder \( p(\boldsymbol{\theta} \mid \mathbf{y}, \mathbf{z}) \) generates the distribution parameters for \(\boldsymbol{\theta}\) based on \(\mathbf{y}\) and \(\mathbf{z}\). Both \(\mathbf{z}\) and \(\boldsymbol{\theta}\) are sampled using the reparameterization trick, ensuring end-to-end differentiability.}
    \label{fig:conditional_vae}
\end{figure}

The key advantage of this formulation lies in the prior distribution \( p(\mathbf{z} \mid \mathbf{y}) \) being conditioned on the observed data \(\mathbf{y}\), allowing the model to adaptively shape its latent representation based on observed data, a crucial feature for effective amortized inference. This adaptive prior serves as a powerful regularization mechanism, ensuring that the latent space accurately captures the relevant features from the joint distribution \( p(\boldsymbol{\theta}, \mathbf{y}) \). As a result, this approach delivers more precise posterior approximations, especially when dealing with data that exhibit complex conditional dependencies, which a fixed prior cannot adequately capture. Such adaptability is crucial for models aiming to generalize across a wide range of queries, making the CP-VAE framework well-suited for real-world applications with intricate data dependencies \citep{NIPS2015_8d55a249}.

The loss function in \eqref{eq:optimization_problem_min} is intractable due to the complexity of computing the KL divergence between the variational distribution \( q_{\psi}(\mathbf{z} \mid \mathbf{y}, \boldsymbol{\theta}) \) and the conditional prior \( p(\mathbf{z} \mid \mathbf{y}) \), as well as the expectation over the latent variable \( \mathbf{z} \). To address this, we approximate \( p(\mathbf{z} \mid \mathbf{y}) \sim \mathcal{N}(\boldsymbol{\mu}_{\mathbf{y}}, \boldsymbol{\Sigma}_{\mathbf{y}}) \), \( q_{\psi}(\mathbf{z} \mid \mathbf{y}, \boldsymbol{\theta}) \sim \mathcal{N}(\boldsymbol{\mu}_{q}, \boldsymbol{\Sigma}_{q}) \), and \( p(\boldsymbol{\theta} \mid \mathbf{y}, \mathbf{z}) \sim \mathcal{N}(\boldsymbol{\mu}_{\theta}, \boldsymbol{\Sigma}_{\theta}) \) as multivariate Gaussians with diagonal covariance matrices, where \( \boldsymbol{\mu}_{\mathbf{y}} \), \( \boldsymbol{\mu}_{q} \), and \( \boldsymbol{\mu}_{\theta} \) are mean vectors, and \( \boldsymbol{\Sigma}_{\mathbf{y}} \), \( \boldsymbol{\Sigma}_{q} \), and \( \boldsymbol{\Sigma}_{\theta} \) are covariance matrices. Here, \( \mathbf{z} \) represents the latent variable with dimensionality \( k_\mathbf{z} \). The final loss is:
\begin{equation}
\begin{aligned}
\mathcal{L}_{\text{CP-VAE}} &= \frac{1}{2} \bigg( \log \frac{|\boldsymbol{\Sigma}_{\mathbf{y}}|}{|\boldsymbol{\Sigma}_{q}|} - k_\mathbf{z} + \mathrm{tr}(\boldsymbol{\Sigma}_{\mathbf{y}}^{-1} \boldsymbol{\Sigma}_{q}) + (\boldsymbol{\mu}_{\mathbf{y}} - \boldsymbol{\mu}_{q})^\top \boldsymbol{\Sigma}_{\mathbf{y}}^{-1} (\boldsymbol{\mu}_{\mathbf{y}} - \boldsymbol{\mu}_{q}) \bigg) \\
&\quad + \sum_{i=1}^{k} \log \sigma_{\theta_i}  + \frac{1}{2N} \sum_{j=1}^{N} \sum_{i=1}^{k} \frac{(\theta_{i,j} - \mu_{\theta_i})^2}{\sigma_{\theta_i}^2},
\end{aligned}
\label{eq:final_loss_function}
\end{equation}

where \( k \) is the dimensionality of \( \boldsymbol{\theta} \). The model architecture is depicted in Figure~\ref{fig:conditional_vae}.

To perform amortized inference with CP-VAE for a new observation \(\mathbf{y}_0\), we first sample \(N\) latent variables \(\{\mathbf{z}^{(i)}\}_{i=1}^N\) from the conditional prior \(p(\mathbf{z} \mid \mathbf{y}_0)\) using the reparameterization trick \citep{kingma2022autoencodingvariationalbayes}. For each latent variable \(\mathbf{z}^{(i)}\), the decoder generates the mean and variance of the parameters, from which \(\boldsymbol{\theta}^{(i)}\) is then sampled via reparameterization. By aggregating a large number of such samples, we can empirically approximate \(p(\boldsymbol{\theta} \mid \mathbf{y}_0)\), enabling statistical analysis and inference of \(\boldsymbol{\theta}\) based on the observed data. The conditional prior, which is dependent on \(\mathbf{y}_0\), allows the model to adapt its latent representation to the specific data, while the generative decoder maps latent variables to the parameter space, enabling CP-VAE to model complex, data-dependent relationships.

\begin{figure}[!t]
    \centering
    \includegraphics[width=5cm]{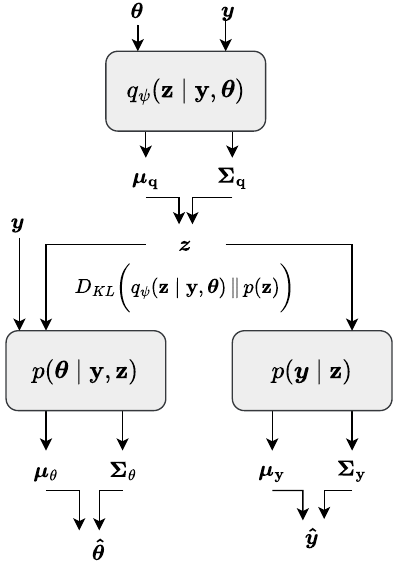}
    \caption{UP-VAE architecture. The encoder \( q_{\psi}(\mathbf{z} \mid \mathbf{y}, \boldsymbol{\theta}) \) outputs the mean and variance of the approximate distribution, from which \( \mathbf{z} \) is sampled (reparameterization trick). The KL divergence is computed between the encoder output and a standard normal prior. The theta decoder \( p(\boldsymbol{\theta} \mid \mathbf{y}, \mathbf{z}) \) outputs the parameters for \(\boldsymbol{\theta}\), and the data decoder \( p(\mathbf{y} \mid \mathbf{z}) \) outputs the parameters for \(\mathbf{y}\) (reparameterization trick).}
    \label{fig:conditional_vae_2}
    \label{fig:conditional_vae_2}
\end{figure}

\subsection{Unconditional Prior VAE (UP-VAE)}
\label{model_b}
A simpler alternative optimization scheme is also proposed, that employs an unconditional prior, in contrast to the earlier approach where the prior was conditioned on the observed data \(\mathbf{y}\). Conditioning the prior on \(\mathbf{y}\) is beneficial for amortized inference, but adds additional complexity to the model as the prior now depends on the data, potentially increasing the computational burden and complicating inference. Moreover, a data-dependent prior can become overly tailored to the training data, reducing the model's ability to generalize to new observations. By utilizing an unconditional prior, we simplify the model architecture and promote better generalization by preventing overfitting to training data. To address this, we use Bayes' theorem to reformulate the conditional distribution \( p(\mathbf{z} \mid \boldsymbol{\theta}, \mathbf{y}) \) as:
\begin{equation}
p(\mathbf{z} \mid \boldsymbol{\theta}, \mathbf{y}) = \frac{p(\boldsymbol{\theta}, \mathbf{y} \mid \mathbf{z}) p(\mathbf{z})}{p(\boldsymbol{\theta}, \mathbf{y})}.
\label{eq:bayes_conditional}
\end{equation}
Substituting \eqref{eq:bayes_conditional} into the KL divergence formulation in \eqref{eq:KL_latent}, and denoting \(D_{KL}(q_{\psi}(\mathbf{z} \mid \mathbf{y}, \boldsymbol{\theta}) \| p(\mathbf{z} \mid \mathbf{y}, \boldsymbol{\theta}))\) as \(\mathcal{D}\),
\begin{equation}
\begin{aligned}
\mathcal{D} &= \mathbb{E}_{\mathbf{z} \sim q_{\psi}} \bigg[\log q_{\psi}(\mathbf{z} \mid \boldsymbol{\theta}, \mathbf{y}) - \log p(\mathbf{z})\bigg] - \mathbb{E}_{\mathbf{z} \sim q_{\psi}} \bigg[\log p(\boldsymbol{\theta}, \mathbf{y} \mid \mathbf{z}) - \log p(\boldsymbol{\theta}, \mathbf{y}) \bigg].
\end{aligned}
\label{eq:kl_divergence_expanded_v3}
\end{equation}
Applying the chain rule to the joint probability \( p(\boldsymbol{\theta}, \mathbf{y}) = p(\mathbf{y}) \, p(\boldsymbol{\theta} \mid \mathbf{y}) \) and recognizing that \( \log p(\mathbf{y}) \) is a constant with respect to \( \boldsymbol{\theta} \), along with the non-negativity property of the KL divergence, we obtain:
\begin{equation}
\begin{aligned}
\log p(\boldsymbol{\theta} \mid \mathbf{y}) &\geq \mathbb{E}_{\mathbf{z} \sim q_{\psi}} \bigg[\log p(\boldsymbol{\theta} \mid \mathbf{y}, \mathbf{z}) + \log p(\mathbf{y} \mid \mathbf{z})\bigg]- D_{KL}\bigg(q_{\psi}(\mathbf{z} \mid \boldsymbol{\theta}, \mathbf{y}) \,\|\, p(\mathbf{z})\bigg).
\end{aligned}
\label{eq:final_inequality}
\end{equation}
Similar to \eqref{eq:elbo}, minimizing the negative of the right-hand side of the derived inequality is equivalent to minimizing the negative log-posterior. This aligns with our primary objective and thus, the final optimization problem can be expressed as:
\begin{equation}
\begin{aligned}
\mathcal{L}(\boldsymbol{\theta}, \mathbf{y}; \psi) = & \, D_{KL}\big(q_{\psi}(\mathbf{z} \mid \boldsymbol{\theta}, \mathbf{y}) \,\|\, p(\mathbf{z})\big) - \mathbb{E}_{\mathbf{z} \sim q_{\psi}} \big[ \log p(\boldsymbol{\theta} \mid \mathbf{y}, \mathbf{z}) \big] - \mathbb{E}_{\mathbf{z} \sim q_{\psi}} \big[ \log p(\mathbf{y} \mid \mathbf{z}) \big].
\end{aligned}
\label{eq:final_optimization}
\end{equation}
This formulation employs two decoder networks: one for reconstructing the parameters \( \boldsymbol{\theta} \) and another for reconstructing the observed data \( \mathbf{y} \). The data decoder introduces an auxiliary loss, complementing the primary loss from the parameter decoder. This auxiliary objective ensures the latent space captures features informative for both \( \boldsymbol{\theta} \) and \( \mathbf{y} \) \citep{10.5555/3666122.3667294}. For amortized inference, only the parameter decoder is required, as it directly generates parameter estimates conditioned on \( \mathbf{y} \). Similar to the previous approach, we approximate the prior \( p(\mathbf{z}) \sim \mathcal{N}(\mathbf{0}, \mathbf{I}) \), the encoder \( q_{\psi}(\mathbf{z} \mid \boldsymbol{\theta}, \mathbf{y}) \sim \mathcal{N}(\boldsymbol{\mu}_{q}, \boldsymbol{\Sigma}_{q}) \), the parameter decoder \( p(\boldsymbol{\theta} \mid \mathbf{y}, \mathbf{z}) \sim \mathcal{N}(\boldsymbol{\mu}_{\theta}, \boldsymbol{\Sigma}_{\theta}) \), and the data decoder \( p(\mathbf{y} \mid \mathbf{z}) \sim \mathcal{N}(\boldsymbol{\mu}_{\mathbf{y}}, \boldsymbol{\Sigma}_{\mathbf{y}}) \) using Gaussian distributions with diagonal covariance matrices, simplifying intractable computations. The model architecture is illustrated in Figure~\ref{fig:conditional_vae_2}. The final loss is:
\begin{equation}
\begin{aligned}
\mathcal{L}_{\text{UP-VAE}} &= \frac{1}{2} \left( -\log |\boldsymbol{\Sigma}_{q}| - k_{\mathbf{z}} + \mathrm{tr}(\boldsymbol{\Sigma}_{q}) + \boldsymbol{\mu}_{q}^\top \boldsymbol{\mu}_{q} \right) + \sum_{i=1}^{k_{\theta}} \log \sigma_{\theta_i} + \frac{1}{2N} \sum_{j=1}^{N} \sum_{i=1}^{k_{\theta}} \frac{(\theta_{i,j} - \mu_{\theta_i})^2}{\sigma_{\theta_i}^2} \\
&\quad + \sum_{i=1}^{k_{\mathbf{y}}} \log \sigma_{\mathbf{y}_i} + \frac{1}{2N} \sum_{j=1}^{N} \sum_{i=1}^{k_{\mathbf{y}}} \frac{(y_{i,j} - \mu_{\mathbf{y}_i})^2}{\sigma_{\mathbf{y}_i}^2},
\end{aligned}
\label{eq:final_loss_function_v2}
\end{equation}

where \( k_{\mathbf{z}} \), \( k_{\theta} \), and \( k_{\mathbf{y}} \) denote the dimensionalities of \( \mathbf{z} \), \( \boldsymbol{\theta} \), and \( \mathbf{y} \), respectively.

Amortized inference using UP-VAE proceeds 
by first sampling \(N\) latent variables \(\{\mathbf{z}^{(i)}\}_{i=1}^N\) from the unconditional prior \( p(\mathbf{z}) \). These latent variables, along with the observed data \(\mathbf{y}_0\), are then input into the trained theta decoder \( p(\boldsymbol{\theta} \mid \mathbf{y}_0, \mathbf{z}^{(i)}) \), which outputs the parameters of the theta distribution from which \(\boldsymbol{\theta}^{(i)}\) are sampled. Aggregating these samples empirically approximates \( p(\boldsymbol{\theta} \mid \mathbf{y}_0) \) in an efficient manner.

\section{Experiments}
\label{Experiments}
We evaluate the proposed approach on ten benchmark problems from the \texttt{sbibm} suite \citep{lueckmann2021benchmarking}, each presenting unique challenges for simulation-based inference. For each test problem, the simulation budgets is varied in \{10,000, 20,000, 30,000\}. We then estimate the posterior distributions by generating 10,000 samples from fixed observed data across five independent runs. To evaluate the accuracy of the posterior approximations, we compare them against reference posteriors obtained via MCMC sampling, using the Maximum Mean Discrepancy (MMD) and the Classifier Two-Sample Test (C2ST) \citep{Friedman:2003id} evaluation metrics. For consistency and reproducibility, we utilized pre-generated observed data and reference posteriors from the \texttt{sbibm} library \citep{lueckmann2021benchmarking}. The C2ST metric computations follow \citep{lueckmann2021benchmarking} -- a C2ST score close to 0.5 indicates that the two distributions are indistinguishable, whereas a score significantly higher than 0.5 suggests notable differences.  The MMD calculation were performed using the BayesFlow framework \citep{bayesflow_2023_software}. All experiments were performed on a machine with 2 $\times$ 16 core Intel(R) Xeon(R) Gold 6226R CPU, 576GB RAM and a NVIDIA Tesla T4 GPU (16 GB RAM).

\subsection{SBI Benchmark Problems}\label{analysis}
This section outlines distinctive challenges posed by each SBI benchmarks. The \textbf{Two Moons} problem tests methods ability to capture bimodal crescent-shaped posteriors. In the \textbf{Gaussian Mixture Model}, the main challenge is identifying the shared mean between two Gaussian distributions with different variances. \textbf{SLCP} introduces complex parameter dependencies (uniform, Gaussian, multimodal, and skewed distributions), while its \textbf{distractors} variant adds uninformative dimensions that obscure inference. \textbf{Gaussian Linear} examines posterior estimation under conjugate priors, whereas \textbf{Gaussian Linear Uniform} extends this challenge to non-conjugate uniform priors. In \textbf{Bernoulli GLM} , the challenge involves estimating posteriors from Bernoulli observations with Gaussian priors with the \textbf{Raw} variant omitting sufficient statistics to increase complexity.  The \textbf{SIR} epidemic model aims to infer contact and recovery rates from noisy infection data over time.  Finally, the \textbf{Lotka-Volterra} model is challenging due to its nonlinear dynamics and potential chaotic behavior.

\begin{figure}[h]
    \centering
    \begin{subfigure}[b]{0.4\textwidth} 
        \centering
        \includegraphics[width=\textwidth]{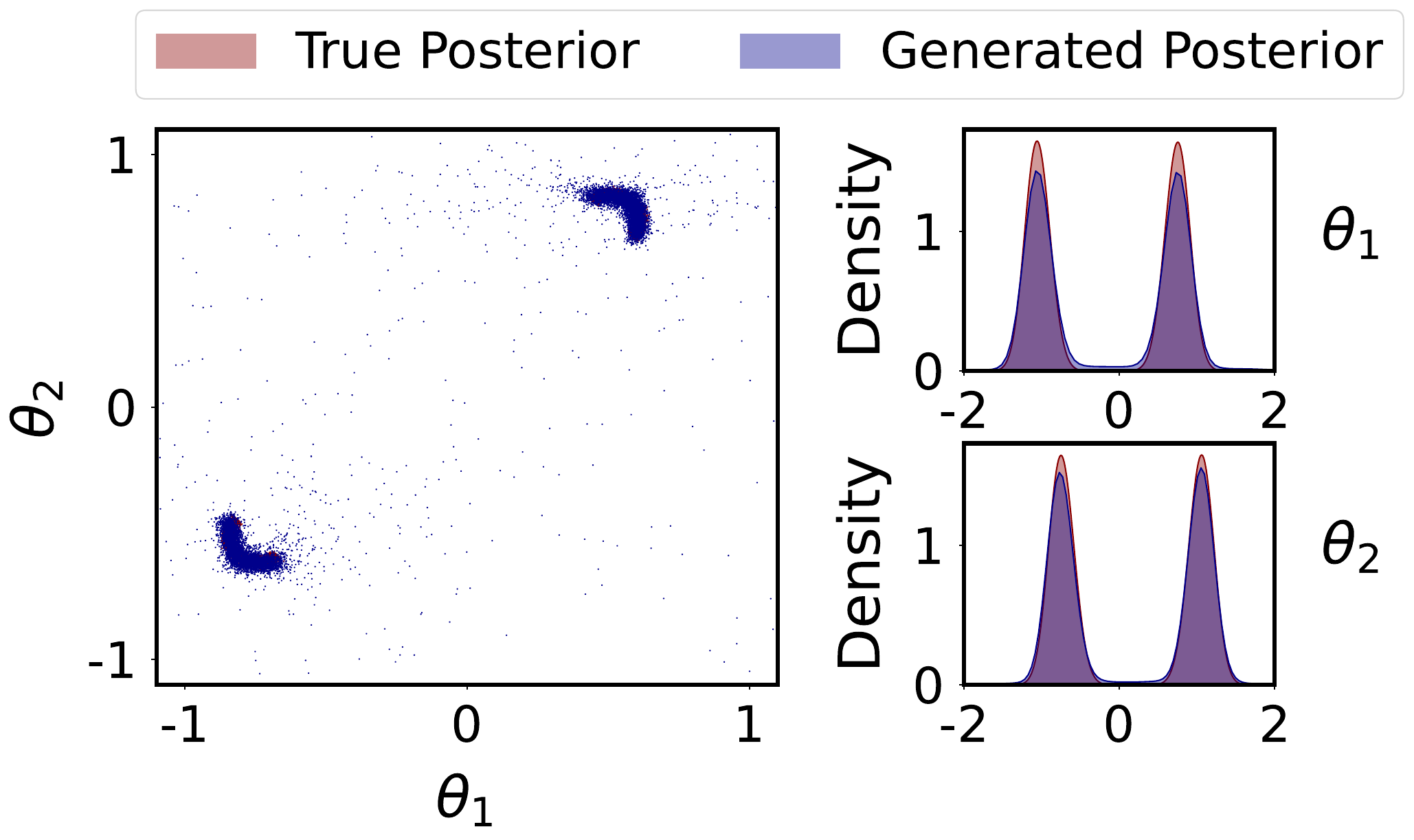}          
        \caption{CP-VAE}
        \label{fig:2moon_v1}
    \end{subfigure}
    \vspace{0.1cm}  
    \begin{subfigure}[b]{0.4\textwidth}
        \centering
        \includegraphics[width=\textwidth]{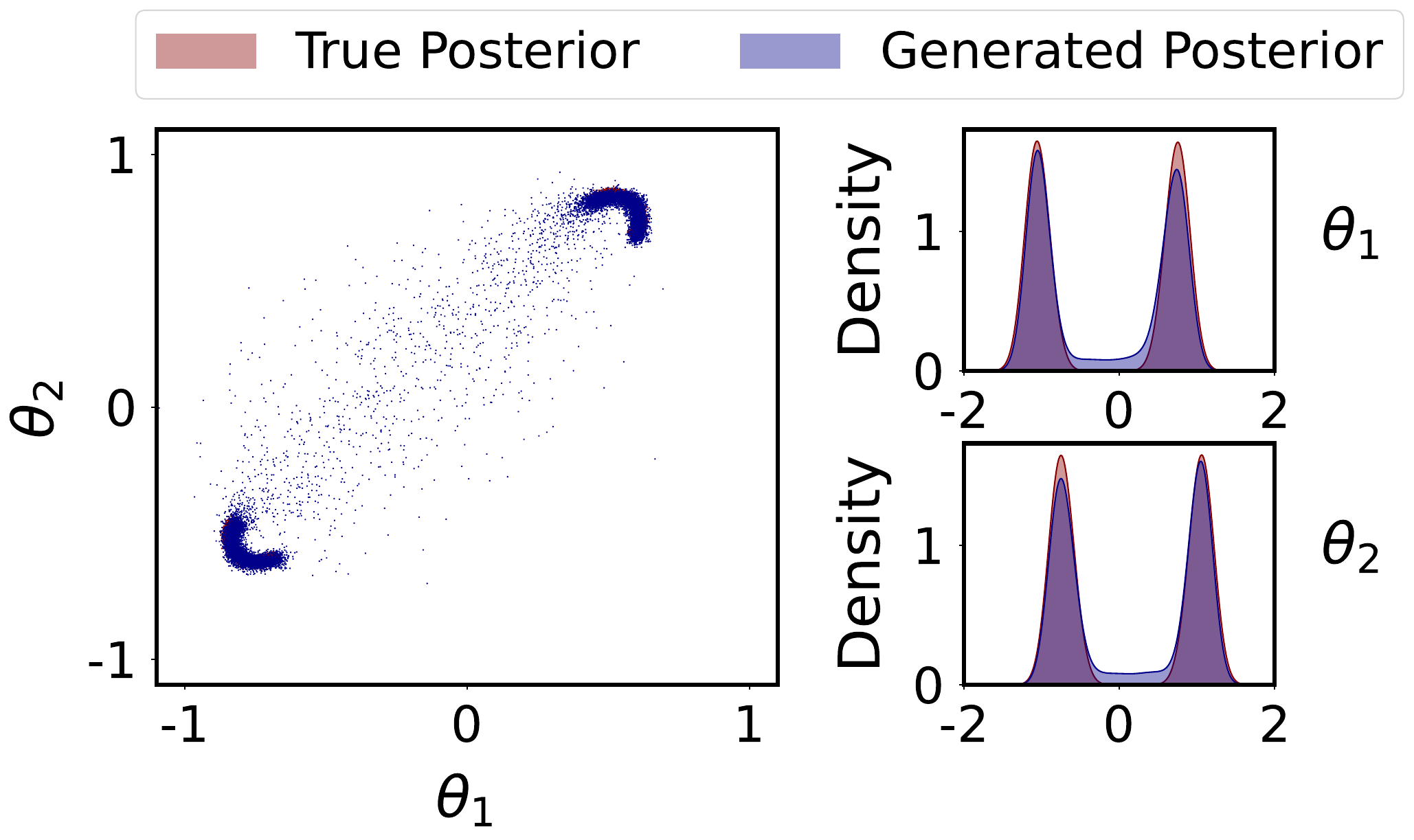}
        \caption{UP-VAE}
        \label{fig:2moon_v2}
    \end{subfigure}
    
    \caption{Two moons: true v/s estimated posterior (sim. budget: 30,000). (a) CP-VAE and (b) UP-VAE.}
    \label{fig:true_posterior_2moons}
\end{figure}

\begin{figure}[h]
    \centering
    \begin{subfigure}[b]{0.5\textwidth}
        \centering
        \includegraphics[width=\textwidth]{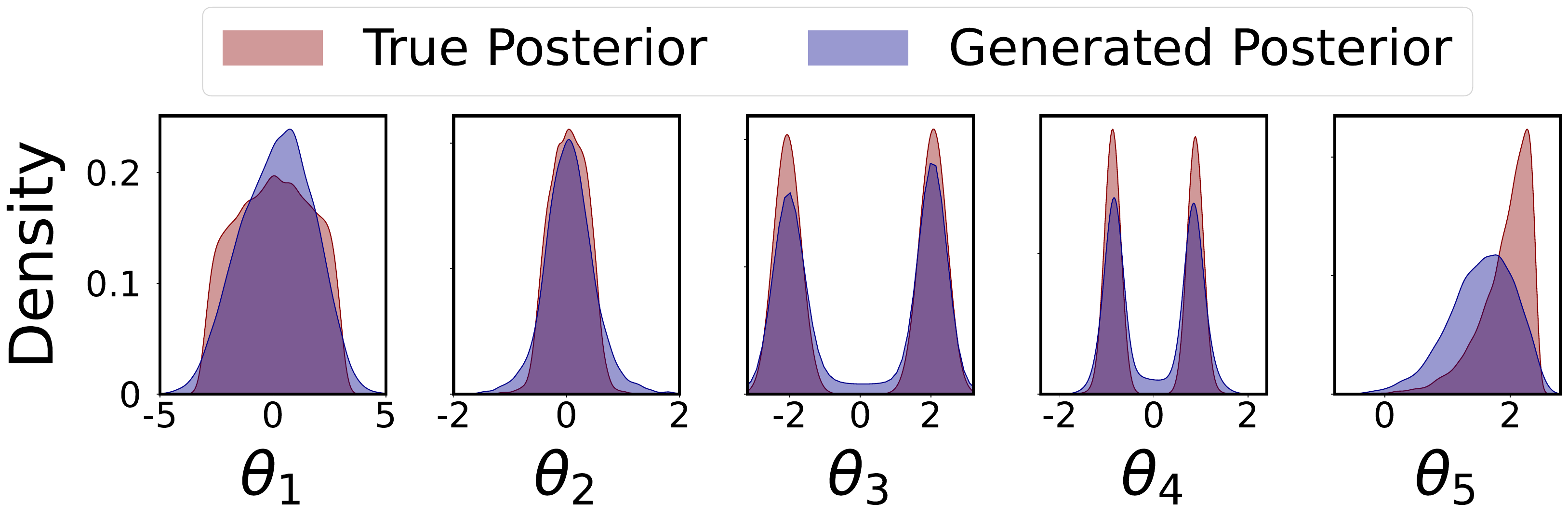}
        \caption{CP-VAE}
        \label{fig:slcp_figure1}
    \end{subfigure}
    
    \vspace{0.1cm} 
    
    \begin{subfigure}[b]{0.5\textwidth}
        \centering
        \includegraphics[width=\textwidth]{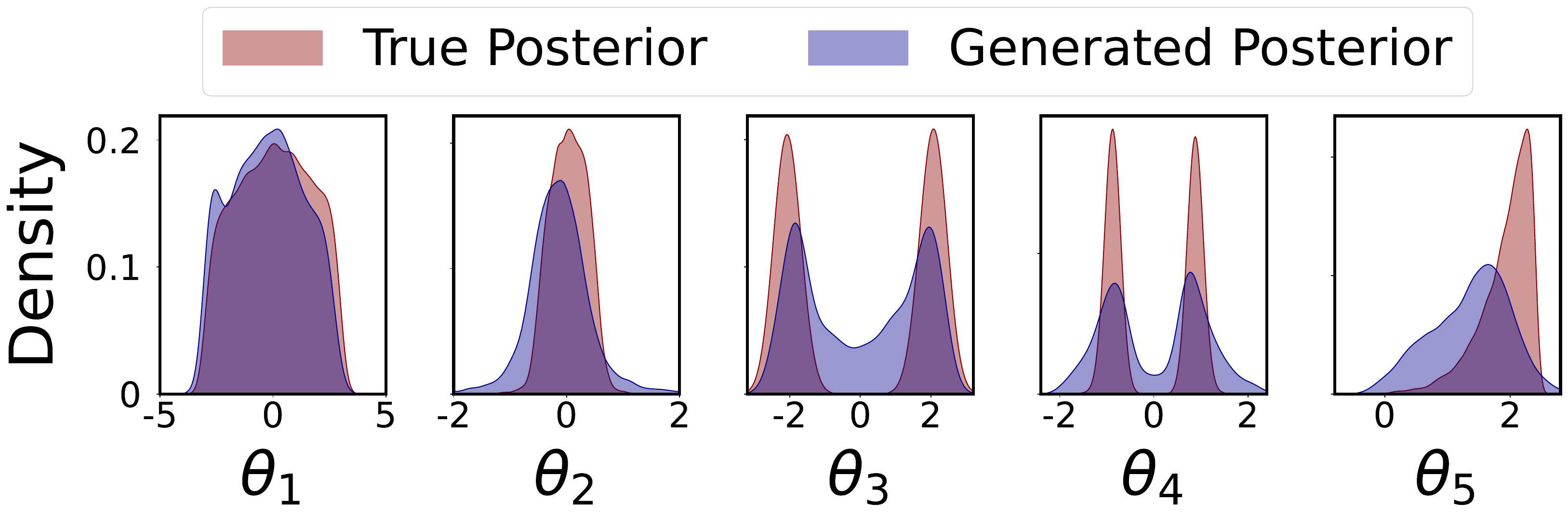}
        \caption{UP-VAE}
        \label{fig:slcp_figure2}
    \end{subfigure}
    \caption{SLCP : true v/s estimated posterior distribution (sim. budget: 30,000). (a) CP-VAE and (b) UP-VAE.}
    \label{fig:slcp_post_v1_v2}
\end{figure}

\subsection{Posterior Visualization on Multimodal Benchmarks}
In this section, we present posterior plots for two challenging benchmark problems with multimodal and distinctive shapes. Figure~\ref{fig:true_posterior_2moons} shows that both CP-VAE and UP-VAE successfully capture the complex shape and multimodal distribution of the data of the two moons model. Additionally, Figure~\ref{fig:slcp_post_v1_v2} demonstrates their ability to model the true posterior's shape across dimensions, despite varying distributions. These results highlight the models' robustness in handling complex posterior structures.

\begin{table*}[!t]
\centering
\caption{C2ST metric for different methods across simulation budgets (mean $\pm$ std. dev.).}
\renewcommand{\arraystretch}{1.25}
\setlength{\tabcolsep}{5pt}
\label{tab:c2st_global}

\begin{subtable}{\textwidth}
\centering
\caption{Two Moons, Bernoulli GLM, and SIR}
\label{tab:c2st_global_1}
\begin{adjustbox}{max width=\textwidth}
\begin{tabular}{l|ccc|ccc|ccc}
\rowcolor{gray!10}
\textbf{Method} 
  & \multicolumn{3}{c|}{\textbf{Two Moons}} 
  & \multicolumn{3}{c|}{\textbf{Bernoulli GLM}} 
  & \multicolumn{3}{c}{\textbf{SIR}} \\

\rowcolor{gray!10}
\textbf{Budget}
  & \textbf{10,000} & \textbf{20,000} & \textbf{30,000}
  & \textbf{10,000} & \textbf{20,000} & \textbf{30,000}
  & \textbf{10,000} & \textbf{20,000} & \textbf{30,000} \\
\hline
GATSBI  
& 0.7770 \tiny{$\pm$0.0630}
& 0.6634 \tiny{$\pm$0.0503}
& 0.7467 \tiny{$\pm$0.1036}
& 0.9999 \tiny{$\pm$0.0001}
& 0.9999 \tiny{$\pm$0.0000}
& 0.9997 \tiny{$\pm$0.0005}
& 0.9990 \tiny{$\pm$0.0015}
& 0.9912 \tiny{$\pm$0.0161}
& 0.9838 \tiny{$\pm$0.0157} \\
NPE
& 0.5749 \tiny{$\pm$0.0331}
& 0.5931 \tiny{$\pm$0.0329}
& 0.5411 \tiny{$\pm$0.0195}
& 0.9121 \tiny{$\pm$0.0338}
& 0.8988 \tiny{$\pm$0.0306}
& 0.8898 \tiny{$\pm$0.0565}
& \textbf{0.8378} \tiny{$\pm$0.0396}
& 0.9122 \tiny{$\pm$0.0207}
& 0.9405 \tiny{$\pm$0.0125} \\
APT
& \textbf{0.5189} \tiny{$\pm$0.0080}
& 0.5202 \tiny{$\pm$0.0114}
& 0.5147 \tiny{$\pm$0.0125}
& 0.8404 \tiny{$\pm$0.0107}
& 0.8328 \tiny{$\pm$0.0156}
& 0.8353 \tiny{$\pm$0.0164}
& 0.9433 \tiny{$\pm$0.0076}
& 0.9434 \tiny{$\pm$0.0061}
& 0.9457 \tiny{$\pm$0.0068} \\
JANA
& 0.5206 \tiny{$\pm$0.0053}
& \textbf{0.5151} \tiny{$\pm$0.0055}
& 0.5123 \tiny{$\pm$0.0032}
& 0.9678 \tiny{$\pm$0.0035}
& 0.9692 \tiny{$\pm$0.0026}
& 0.9589 \tiny{$\pm$0.0034}
& 0.8697 \tiny{$\pm$0.0149}
& 0.9175 \tiny{$\pm$0.0059}
& 0.9343 \tiny{$\pm$0.0055} \\
Simformer
& 0.5273 \tiny{$\pm$0.0229}
& 0.5190 \tiny{$\pm$0.0174}
& \textbf{0.5098} \tiny{$\pm$0.0084}
& \textbf{0.6584} \tiny{$\pm$0.0381}
& \textbf{0.6131} \tiny{$\pm$0.0225}
& \textbf{0.6154} \tiny{$\pm$0.0265}
& 0.9378 \tiny{$\pm$0.0053}  
& 0.9520 \tiny{$\pm$0.0064} 
& 0.9478 \tiny{$\pm$0.0049} \\

\rowcolor{blue!5}
CP-VAE
& 0.6443 \tiny{$\pm$0.0288}
& 0.6072 \tiny{$\pm$0.0108}
& 0.6346 \tiny{$\pm$0.0303}
& 0.6955 \tiny{$\pm$0.0239}
& 0.6697 \tiny{$\pm$0.0361}
& 0.6427 \tiny{$\pm$0.0177}
& 0.8524 \tiny{$\pm$0.0719}
& 0.8976 \tiny{$\pm$0.0121}
& 0.8857 \tiny{$\pm$0.0330} \\
\rowcolor{blue!5}
UP-VAE
& 0.6510 \tiny{$\pm$0.0310}
& 0.6565 \tiny{$\pm$0.0364}
& 0.6212 \tiny{$\pm$0.0252}
& 0.8270 \tiny{$\pm$0.0249}
& 0.8311 \tiny{$\pm$0.0216}
& 0.7819 \tiny{$\pm$0.0692}
& 0.8390 \tiny{$\pm$0.0378}
& \textbf{0.8731} \tiny{$\pm$0.0482}
& \textbf{0.8893} \tiny{$\pm$0.0317} \\
\end{tabular}
\end{adjustbox}
\end{subtable}

\vspace{0.5cm} 

\begin{subtable}{\textwidth}
\centering
\caption{SLCP Distractors, Bernoulli GLM Raw, and Lotka Volterra}
\label{tab:c2st_global_2}
\begin{adjustbox}{max width=\textwidth}
\begin{tabular}{l|ccc|ccc|ccc}
\rowcolor{gray!10}
\textbf{Method} 
  & \multicolumn{3}{c|}{\textbf{SLCP Distractors}} 
  & \multicolumn{3}{c|}{\textbf{Bernoulli GLM Raw}} 
  & \multicolumn{3}{c}{\textbf{Lotka Volterra}} \\
  
\rowcolor{gray!10}
\textbf{Budget}
  & \textbf{10,000} & \textbf{20,000} & \textbf{30,000}
  & \textbf{10,000} & \textbf{20,000} & \textbf{30,000}
  & \textbf{10,000} & \textbf{20,000} & \textbf{30,000} \\
\hline
GATSBI
& 0.9995 \tiny{$\pm$0.0007}
& 0.9978 \tiny{$\pm$0.0016}
& 0.9994 \tiny{$\pm$0.0004}
& 0.9998 \tiny{$\pm$0.0001}
& 0.9999 \tiny{$\pm$0.0001}
& 0.9999 \tiny{$\pm$0.0000}
& 0.9999 \tiny{$\pm$0.0001}
& 0.9999 \tiny{$\pm$0.0000}
& 1.0000 \tiny{$\pm$0.0000} \\
NPE
& 0.9714 \tiny{$\pm$0.0144}
& 0.9468 \tiny{$\pm$0.0157}
& 0.9359 \tiny{$\pm$0.0127}
& 0.9354 \tiny{$\pm$0.0093}
& 0.9114 \tiny{$\pm$0.0082}
& 0.9065 \tiny{$\pm$0.0105}
& 0.9999 \tiny{$\pm$0.0001}
& 0.9999 \tiny{$\pm$0.0000}
& 0.9999 \tiny{$\pm$0.0000} \\
JANA
& 0.9987 \tiny{$\pm$0.0010}
& 0.9945 \tiny{$\pm$0.0138}
& 0.9933 \tiny{$\pm$0.0654}
& 0.9997 \tiny{$\pm$0.0007}
& 1.0000 \tiny{$\pm$0.0000}
& 1.0000 \tiny{$\pm$0.0000}
& 0.9999 \tiny{$\pm$0.0000}
& 0.9998 \tiny{$\pm$0.0001}
& 0.9998 \tiny{$\pm$0.0001} \\
APT
& 0.9823 \tiny{$\pm$0.0055}
& 0.9451 \tiny{$\pm$0.0158}
& 0.9130 \tiny{$\pm$0.0301}
& \textbf{0.8455} \tiny{$\pm$0.0116}
& \textbf{0.8596} \tiny{$\pm$0.0152}
& 0.8461 \tiny{$\pm$0.0135}
& 0.9999 \tiny{$\pm$0.0001}
& 0.9999 \tiny{$\pm$0.0000}
& 0.9999 \tiny{$\pm$0.0001} \\
Simformer
& \textbf{0.9198} \tiny{$\pm$0.0451}
& \textbf{0.8814} \tiny{$\pm$0.0433}
& \textbf{0.8717} \tiny{$\pm$0.0453}
& 0.9814 \tiny{$\pm$0.0106}
& 0.8600 \tiny{$\pm$0.0452}
& \textbf{0.7750} \tiny{$\pm$0.0473}
& 0.9999 \tiny{$\pm$0.0003} 
& 0.9999 \tiny{$\pm$0.0012} 
& 1.000 \tiny{$\pm$0.0002} \\
\rowcolor{blue!5}
CP-VAE
& 0.9956 \tiny{$\pm$0.0037}
& 0.9880 \tiny{$\pm$0.0057}

& 0.9746 \tiny{$\pm$0.0075}
& 0.9491 \tiny{$\pm$0.0078}
& 0.9171 \tiny{$\pm$0.0258}
& 0.9307 \tiny{$\pm$0.0000}
& 0.9999 \tiny{$\pm$0.0001}
& 0.9999 \tiny{$\pm$0.0001}
& 0.9999 \tiny{$\pm$0.0001} \\
\rowcolor{blue!5}
UP-VAE
& 0.9838 \tiny{$\pm$0.0036}
& 0.9798 \tiny{$\pm$0.0023}
& 0.9840 \tiny{$\pm$0.0053}
& 0.8674 \tiny{$\pm$0.0493}
& 0.8790 \tiny{$\pm$0.0363}
& 0.8536 \tiny{$\pm$0.0460}
& 0.9999 \tiny{$\pm$0.0001}
& 0.9999 \tiny{$\pm$0.0001}
& 0.9999 \tiny{$\pm$0.0001} \\
\end{tabular}
\end{adjustbox}
\end{subtable}

\vspace{0.5cm} 

\begin{subtable}{\textwidth}
\centering
\caption{Gaussian Mixture, Gaussian Linear, Gaussian Linear Uniform, and SLCP}
\label{tab:c2st_global_3}
\renewcommand{\arraystretch}{1.5}
\setlength{\tabcolsep}{2pt}
\begin{adjustbox}{max width=\textwidth}
\begin{tabular}{l|ccc|ccc|ccc|ccc}
\rowcolor{gray!10}
\textbf{Method}
  & \multicolumn{3}{c|}{\textbf{Gaussian Mixture}}
  & \multicolumn{3}{c|}{\textbf{Gaussian Linear}}
  & \multicolumn{3}{c|}{\textbf{Gaussian Linear Uniform}}
  & \multicolumn{3}{c}{\textbf{SLCP}} \\

\rowcolor{gray!10}
\textbf{Budget}
  & \textbf{10,000} & \textbf{20,000} & \textbf{30,000}
  & \textbf{10,000} & \textbf{20,000} & \textbf{30,000}
  & \textbf{10,000} & \textbf{20,000} & \textbf{30,000}
  & \textbf{10,000} & \textbf{20,000} & \textbf{30,000} \\
\hline
GATSBI
& 0.7474 \tiny{$\pm$0.0364}
& 0.6977 \tiny{$\pm$0.0260}
& 0.7234 \tiny{$\pm$0.0599}
& 0.9930 \tiny{$\pm$0.0038}
& 0.9916 \tiny{$\pm$0.0046}
& 0.9929 \tiny{$\pm$0.0059}
& 0.9989 \tiny{$\pm$0.0004}
& 0.9976 \tiny{$\pm$0.0016}
& 0.9988 \tiny{$\pm$0.0010}
& 0.9687 \tiny{$\pm$0.0286}
& 0.9436 \tiny{$\pm$0.0203}
& 0.9723 \tiny{$\pm$0.0140} \\
NPE
& 0.6118 \tiny{$\pm$0.0155}
& 0.5830 \tiny{$\pm$0.0240}
& 0.5818 \tiny{$\pm$0.0158}
& 0.5472 \tiny{$\pm$0.0076}
& \textbf{0.5238} \tiny{$\pm$0.0112}
& 0.5266 \tiny{$\pm$0.0037}
& 0.5562 \tiny{$\pm$0.0147}
& \textbf{0.5292} \tiny{$\pm$0.0068}
& \textbf{0.5256} \tiny{$\pm$0.0066}
& 0.8996 \tiny{$\pm$0.0206}
& 0.8502 \tiny{$\pm$0.0174}
& 0.8546 \tiny{$\pm$0.0213} \\
JANA
& 0.6146 \tiny{$\pm$0.0066}
& 0.5955 \tiny{$\pm$0.0026}
& 0.5854 \tiny{$\pm$0.0037}
& 0.9905 \tiny{$\pm$0.0028}
& 0.9903 \tiny{$\pm$0.0003}
& 0.9904 \tiny{$\pm$0.0009}
& 0.9845 \tiny{$\pm$0.0027}
& 0.9854 \tiny{$\pm$0.0021}
& 0.9879 \tiny{$\pm$0.0011}
& 0.8707 \tiny{$\pm$0.0345}
& 0.8421 \tiny{$\pm$0.0292}
& 0.7791 \tiny{$\pm$0.0232} \\
APT
& 0.5457 \tiny{$\pm$0.0058}
& 0.5425 \tiny{$\pm$0.0124}
& 0.5467 \tiny{$\pm$0.0050}
& \textbf{0.5367} \tiny{$\pm$0.0038}
& 0.5293 \tiny{$\pm$0.0052}
& \textbf{0.5258} \tiny{$\pm$0.0047}
& \textbf{0.5401} \tiny{$\pm$0.0115}
& 0.5302 \tiny{$\pm$0.0069}
& 0.5264 \tiny{$\pm$0.0075}
& 0.8164 \tiny{$\pm$0.0295}
& 0.6882 \tiny{$\pm$0.0342}
& \textbf{0.6473} \tiny{$\pm$0.0343} \\
Simformer
& \textbf{0.5166} \tiny{$\pm$0.0210}
& \textbf{0.5100} \tiny{$\pm$0.0090}
& \textbf{0.5126} \tiny{$\pm$0.0070}
& 0.6130 \tiny{$\pm$0.0281}
& 0.6226 \tiny{$\pm$0.0398}
& 0.5866 \tiny{$\pm$0.0430}
& 0.6501 \tiny{$\pm$0.0519}
& 0.6207 \tiny{$\pm$0.0519}
& 0.5780 \tiny{$\pm$0.0211}
& \textbf{0.7661} \tiny{$\pm$0.0685}
& \textbf{0.6201} \tiny{$\pm$0.0490}
& 0.7192 \tiny{$\pm$0.0721} \\
\rowcolor{blue!5}
CP-VAE
& 0.7328 \tiny{$\pm$0.0347}
& 0.7492 \tiny{$\pm$0.0496}
& 0.7482 \tiny{$\pm$0.0228}
& 0.6309 \tiny{$\pm$0.0357}
& 0.6003 \tiny{$\pm$0.0165}
& 0.5851 \tiny{$\pm$0.0355}
& 0.7854 \tiny{$\pm$0.0269}
& 0.7169 \tiny{$\pm$0.0332}
& 0.6886 \tiny{$\pm$0.0637}
& 0.9279 \tiny{$\pm$0.0040}
& 0.8987 \tiny{$\pm$0.0164}
& 0.8582 \tiny{$\pm$0.0148} \\
\rowcolor{blue!5}
UP-VAE
& 0.7977 \tiny{$\pm$0.0390}
& 0.8134 \tiny{$\pm$0.0395}
& 0.8032 \tiny{$\pm$0.0418}
& 0.7704 \tiny{$\pm$0.0461}
& 0.6037 \tiny{$\pm$0.0318}
& 0.7530 \tiny{$\pm$0.0060}
& 0.8024 \tiny{$\pm$0.0156}
& 0.7898 \tiny{$\pm$0.0137}
& 0.7549 \tiny{$\pm$0.0276}
& 0.9578 \tiny{$\pm$0.0168}
& 0.9417 \tiny{$\pm$0.0114}
& 0.9409 \tiny{$\pm$0.0162} \\
\bottomrule
\end{tabular}
\end{adjustbox}
\end{subtable}

\end{table*}
 
\begin{table*}[h!]
\centering
\caption{MMD metric for different methods across simulation budgets (mean $\pm$ std. dev.).}
\label{tab:mmd_global}
\renewcommand{\arraystretch}{1.25}
\setlength{\tabcolsep}{5pt}

\begin{subtable}[t]{\textwidth}
\centering
\caption{Two Moons, Bernoulli GLM, and SIR}
\label{tab:mmd_global_1}
\begin{adjustbox}{max width=\textwidth}
\begin{tabular}{l|ccc|ccc|ccc}
\rowcolor{gray!10}
\textbf{Method}
  & \multicolumn{3}{c|}{\textbf{Two Moons}}
  & \multicolumn{3}{c|}{\textbf{Bernoulli GLM}}
  & \multicolumn{3}{c}{\textbf{SIR}} \\

\rowcolor{gray!10}
\textbf{Budget}
  & \textbf{10,000} & \textbf{20,000} & \textbf{30,000}
  & \textbf{10,000} & \textbf{20,000} & \textbf{30,000}
  & \textbf{10,000} & \textbf{20,000} & \textbf{30,000} \\
\hline
GATSBI
& 0.1213 \tiny{$\pm$0.0650}
& 0.0582 \tiny{$\pm$0.0122}
& 0.1432 \tiny{$\pm$0.1161}
& 3.6381 \tiny{$\pm$0.5428}
& 4.8299 \tiny{$\pm$0.6825}
& 3.9965 \tiny{$\pm$1.8510}
& 2.9470 \tiny{$\pm$0.4891}
& 2.7502 \tiny{$\pm$0.7268}
& 2.1466 \tiny{$\pm$0.5625} \\
NPE
& 0.0254 \tiny{$\pm$0.0025}
& 0.0241 \tiny{$\pm$0.0099}
& 0.0125 \tiny{$\pm$0.0045}
& 1.0371 \tiny{$\pm$0.1830}
& 0.9379 \tiny{$\pm$0.1001}
& 0.8675 \tiny{$\pm$0.1945}
& 0.6611 \tiny{$\pm$0.2236}
& 1.0312 \tiny{$\pm$0.1255}
& 1.1903 \tiny{$\pm$0.1007} \\
JANA
& 0.0060 \tiny{$\pm$0.0013}
& 0.0068 \tiny{$\pm$0.0017}
& 0.0047 \tiny{$\pm$0.0009}
& 3.3232 \tiny{$\pm$0.0932}
& 3.1488 \tiny{$\pm$0.0280}
& 2.9010 \tiny{$\pm$0.0878}
& 0.8495 \tiny{$\pm$0.1254}
& 1.1077 \tiny{$\pm$0.0336}
& 1.1421 \tiny{$\pm$0.0392} \\
APT
& 0.0073 \tiny{$\pm$0.0030}
& 0.0069 \tiny{$\pm$0.0030}
& 0.0037 \tiny{$\pm$0.0012}
& 0.7383 \tiny{$\pm$0.0853}
& 0.4944 \tiny{$\pm$0.0663}
& 0.5808 \tiny{$\pm$0.0539}
& 1.1003 \tiny{$\pm$0.0712}
& 1.1093 \tiny{$\pm$0.0311}
& 1.1291 \tiny{$\pm$0.0481} \\
Simformer
& \textbf{0.0005} \tiny{$\pm$0.0003}
& \textbf{0.0001} \tiny{$\pm$0.0001}
& \textbf{0.0003} \tiny{$\pm$0.0004}
& \textbf{0.0191} \tiny{$\pm$0.0098}
& \textbf{0.0156} \tiny{$\pm$0.0087}
& \textbf{0.0116} \tiny{$\pm$0.0061}
& \textbf{0.0018} \tiny{$\pm$0.0028}
& \textbf{0.0016} \tiny{$\pm$0.0042}
& \textbf{0.0016} \tiny{$\pm$0.0085} \\

\rowcolor{blue!5}
CP-VAE
& 0.0382 \tiny{$\pm$0.0110}
& 0.0369 \tiny{$\pm$0.0088}
& 0.0276 \tiny{$\pm$0.0050}
& 0.1966 \tiny{$\pm$0.1375}
& 0.1557 \tiny{$\pm$0.0663}
& 0.0942 \tiny{$\pm$0.0547}
& 0.8792 \tiny{$\pm$0.3327}
& 1.0158 \tiny{$\pm$0.0566}
& 0.9990 \tiny{$\pm$0.1923} \\
\rowcolor{blue!5}
UP-VAE
& 0.0736 \tiny{$\pm$0.0097}
& 0.0795 \tiny{$\pm$0.0151}
& 0.0544 \tiny{$\pm$0.0150}
& 1.5420 \tiny{$\pm$0.4930}
& 1.4748 \tiny{$\pm$0.2623}
& 1.1160 \tiny{$\pm$0.7496}
& 0.5168 \tiny{$\pm$0.1193}
& 0.6884 \tiny{$\pm$0.2044}
& 0.6487 \tiny{$\pm$0.1087} \\
\end{tabular}
\end{adjustbox}
\end{subtable}

\vspace{0.3cm} 

\begin{subtable}[t]{\textwidth}
\centering
\caption{SLCP Distractors, Bernoulli GLM Raw, and Lotka Volterra}
\label{tab:mmd_global_2}
\begin{adjustbox}{max width=\textwidth}
\begin{tabular}{l|ccc|ccc|ccc}
\rowcolor{gray!10}
\textbf{Method}
  & \multicolumn{3}{c|}{\textbf{SLCP Distractors}}
  & \multicolumn{3}{c|}{\textbf{Bernoulli GLM Raw}}
  & \multicolumn{3}{c}{\textbf{Lotka Volterra}} \\

\rowcolor{gray!10}
\textbf{Budget}
  & \textbf{10,000} & \textbf{20,000} & \textbf{30,000}
  & \textbf{10,000} & \textbf{20,000} & \textbf{30,000}
  & \textbf{10,000} & \textbf{20,000} & \textbf{30,000} \\
\hline
GATSBI
& 5.1993 \tiny{$\pm$1.2320}
& 7.2478 \tiny{$\pm$1.2260}
& 3.0631 \tiny{$\pm$0.3244}
& 3.4369 \tiny{$\pm$0.8517}
& 4.3479 \tiny{$\pm$0.5433}
& 3.5950 \tiny{$\pm$0.8613}
& 7.2684 \tiny{$\pm$0.5733}
& 5.5183 \tiny{$\pm$1.7126}
& 4.8858 \tiny{$\pm$1.2146} \\
NPE
& 1.2167 \tiny{$\pm$0.2397}
& 0.7834 \tiny{$\pm$0.3453}
& 0.6428 \tiny{$\pm$0.1779}
& 1.1227 \tiny{$\pm$0.0987}
& 1.1389 \tiny{$\pm$0.1878}
& 1.1448 \tiny{$\pm$0.1287}
& 2.6357 \tiny{$\pm$0.3468}
& 3.8635 \tiny{$\pm$0.2770}
& 4.3310 \tiny{$\pm$0.0709} \\
JANA
& 3.4834 \tiny{$\pm$1.0817}
& 3.1234 \tiny{$\pm$0.1769}
& 3.0176 \tiny{$\pm$0.8765}
& 7.6642 \tiny{$\pm$1.5789}
& 10.4712 \tiny{$\pm$0.9031}
& 11.3721 \tiny{$\pm$0.2927}
& 4.8826 \tiny{$\pm$0.8294}
& 3.7653 \tiny{$\pm$0.4288}
& 3.8278 \tiny{$\pm$0.5220} \\
APT
& 1.8380 \tiny{$\pm$0.4448}
& 0.8278 \tiny{$\pm$0.2373}
& 0.4422 \tiny{$\pm$0.1428}
& 0.7233 \tiny{$\pm$0.2131}
& 0.6968 \tiny{$\pm$0.0826}
& 0.5437 \tiny{$\pm$0.1333}
& 3.1921 \tiny{$\pm$0.8198}
& 3.2722 \tiny{$\pm$0.8169}
& 4.4739 \tiny{$\pm$0.9642} \\
Simformer
& \textbf{0.0540} \tiny{$\pm$0.0173}
& \textbf{0.0541} \tiny{$\pm$0.0161}
& \textbf{0.0552} \tiny{$\pm$0.0195}
& \textbf{0.2056} \tiny{$\pm$0.0486}
& \textbf{0.0926} \tiny{$\pm$0.0493}
& \textbf{0.0731} \tiny{$\pm$0.0238}
& \textbf{0.9696} \tiny{$\pm$0.4434}
& \textbf{0.8342} \tiny{$\pm$0.4742}
& \textbf{0.9634} \tiny{$\pm$0.2588} \\

\rowcolor{blue!5}
CP-VAE
& 3.8511 \tiny{$\pm$1.4076}
& 1.8129 \tiny{$\pm$0.6628}
& 1.2052 \tiny{$\pm$0.6557}
& 2.1708 \tiny{$\pm$0.0391}
& 1.6792 \tiny{$\pm$0.3563}
& 1.6827 \tiny{$\pm$0.0000}
& 3.1960 \tiny{$\pm$0.3824}
& 3.2443 \tiny{$\pm$0.2162}
& 3.4280 \tiny{$\pm$0.2793} \\
\rowcolor{blue!5}
UP-VAE
& 1.9762 \tiny{$\pm$0.3219}
& 1.5244 \tiny{$\pm$0.1978}
& 1.4353 \tiny{$\pm$0.1306}
& 1.2034 \tiny{$\pm$0.6426}
& 1.6952 \tiny{$\pm$1.0074}
& 1.1553 \tiny{$\pm$0.8829}
& 4.4367 \tiny{$\pm$0.7106}
& 6.1012 \tiny{$\pm$0.5511}
& 6.0285 \tiny{$\pm$1.1408} \\
\end{tabular}
\end{adjustbox}
\end{subtable}
\end{table*}


\begin{table*}[!t]
\ContinuedFloat
\centering
\renewcommand{\arraystretch}{1.25}
\setlength{\tabcolsep}{5pt}
\begin{subtable}[t]{\textwidth}
\centering
\caption{Gaussian Mixture, Gaussian Linear, Gaussian Linear Uniform, and SLCP}
\renewcommand{\arraystretch}{1.5}
\setlength{\tabcolsep}{2pt}
\label{tab:mmd_global_3}
\begin{adjustbox}{max width=\textwidth}
\begin{tabular}{l|ccc|ccc|ccc|ccc}
\rowcolor{gray!10}
\textbf{Method}
  & \multicolumn{3}{c|}{\textbf{Gaussian Mixture}}
  & \multicolumn{3}{c|}{\textbf{Gaussian Linear}}
  & \multicolumn{3}{c|}{\textbf{Gaussian Linear Uniform}}
  & \multicolumn{3}{c}{\textbf{SLCP}} \\

\rowcolor{gray!10}
\textbf{Budget}
  & \textbf{10,000} & \textbf{20,000} & \textbf{30,000}
  & \textbf{10,000} & \textbf{20,000} & \textbf{30,000}
  & \textbf{10,000} & \textbf{20,000} & \textbf{30,000}
  & \textbf{10,000} & \textbf{20,000} & \textbf{30,000} \\
\hline
GATSBI
& 0.2242 \tiny{$\pm$0.0296}
& 0.1149 \tiny{$\pm$0.0549}
& 0.1733 \tiny{$\pm$0.1082}
& 0.4999 \tiny{$\pm$0.1595}
& 0.6476 \tiny{$\pm$0.3050}
& 0.7182 \tiny{$\pm$0.2839}
& 0.9044 \tiny{$\pm$0.2353}
& 0.7712 \tiny{$\pm$0.1483}
& 1.1197 \tiny{$\pm$0.4795}
& 0.7197 \tiny{$\pm$0.5598}
& 0.3375 \tiny{$\pm$0.1473}
& 0.7948 \tiny{$\pm$0.5992} \\
NPE
& 0.0604 \tiny{$\pm$0.0260}
& 0.0418 \tiny{$\pm$0.0211}
& 0.0554 \tiny{$\pm$0.0347}
& 0.0105 \tiny{$\pm$0.0036}
& 0.0074 \tiny{$\pm$0.0018}
& 0.0082 \tiny{$\pm$0.0011}
& 0.0258 \tiny{$\pm$0.0098}
& 0.0117 \tiny{$\pm$0.0041}
& 0.0121 \tiny{$\pm$0.0051}
& 0.4021 \tiny{$\pm$0.0551}
& 0.1632 \tiny{$\pm$0.0684}
& 0.2333 \tiny{$\pm$0.0935} \\
JANA
& 0.0579 \tiny{$\pm$0.0079}
& 0.0415 \tiny{$\pm$0.0037}
& 0.0383 \tiny{$\pm$0.0042}
& 0.5375 \tiny{$\pm$0.0367}
& 0.5245 \tiny{$\pm$0.0072}
& 0.5169 \tiny{$\pm$0.0028}
& 0.5798 \tiny{$\pm$0.0357}
& 0.5875 \tiny{$\pm$0.0279}
& 0.6181 \tiny{$\pm$0.0128}
& 0.3673 \tiny{$\pm$0.2011}
& 0.2045 \tiny{$\pm$0.0917}
& 0.1209 \tiny{$\pm$0.0456} \\
APT
& 0.0137 \tiny{$\pm$0.0031}
& 0.0108 \tiny{$\pm$0.0088}
& 0.0110 \tiny{$\pm$0.0034}
& \textbf{0.0075} \tiny{$\pm$0.0013}
& 0.0070 \tiny{$\pm$0.0018}
& 0.0080 \tiny{$\pm$0.0008}
& 0.0166 \tiny{$\pm$0.0087}
& 0.0138 \tiny{$\pm$0.0053}
& 0.0113 \tiny{$\pm$0.0039}
& 0.1405 \tiny{$\pm$0.0347}
& 0.0487 \tiny{$\pm$0.0296}
& 0.0287 \tiny{$\pm$0.0166} \\

Simformer
& \textbf{0.0009} \tiny{$\pm$0.0007}
& \textbf{0.0009} \tiny{$\pm$0.0006}
& \textbf{0.0007} \tiny{$\pm$0.0010}
& 0.0092 \tiny{$\pm$0.0116}
& \textbf{0.0037} \tiny{$\pm$0.0035}
& \textbf{0.0037} \tiny{$\pm$0.0028}
& \textbf{0.0014} \tiny{$\pm$0.0006}
& \textbf{0.0015} \tiny{$\pm$0.0011}
& \textbf{0.0008} \tiny{$\pm$0.0003}
& \textbf{0.0323} \tiny{$\pm$0.0203}
& \textbf{0.0025} \tiny{$\pm$0.0034}
& \textbf{0.0021} \tiny{$\pm$0.0030} \\

\rowcolor{blue!5}
CP-VAE
& 0.1846 \tiny{$\pm$0.0602}
& 0.2623 \tiny{$\pm$0.1369}
& 0.2209 \tiny{$\pm$0.0587}
& 0.0382 \tiny{$\pm$0.0131}
& 0.0220 \tiny{$\pm$0.0058}
& 0.0216 \tiny{$\pm$0.0081}
& 0.0829 \tiny{$\pm$0.0295}
& 0.0448 \tiny{$\pm$0.0070}
& 0.0363 \tiny{$\pm$0.0130}
& 0.4309 \tiny{$\pm$0.1369}
& 0.2488 \tiny{$\pm$0.0557}
& 0.1658 \tiny{$\pm$0.0881} \\
\rowcolor{blue!5}
UP-VAE
& 0.3081 \tiny{$\pm$0.1534}
& 0.3927 \tiny{$\pm$0.1208}
& 0.3684 \tiny{$\pm$0.0876}
& 0.0611 \tiny{$\pm$0.0207}
& 0.0121 \tiny{$\pm$0.0036}
& 0.0446 \tiny{$\pm$0.0025}
& 0.0912 \tiny{$\pm$0.0143}
& 0.1315 \tiny{$\pm$0.0293}
& 0.1246 \tiny{$\pm$0.0187}
& 0.4740 \tiny{$\pm$0.1183}
& 0.4042 \tiny{$\pm$0.1181}
& 0.3946 \tiny{$\pm$0.1852} \\
\end{tabular}
\end{adjustbox}
\end{subtable}
\end{table*}

\begin{table}[h]
\centering
\caption{Training time (in minutes) for different methods across simulation budgets (mean $\pm$ std.\ dev.).}
\renewcommand{\arraystretch}{1.1}
\setlength{\tabcolsep}{5pt}
\label{tab:train_times_all}

\begin{subtable}[t]{\textwidth}
\centering
\caption{Two Moons, Bernoulli GLM, SIR.}
\begin{adjustbox}{max width=\textwidth}
\begin{tabular}{l|ccc|ccc|ccc}
\rowcolor{gray!10}
\textbf{Method}
  & \multicolumn{3}{c|}{\textbf{Two Moons}}
  & \multicolumn{3}{c|}{\textbf{Bernoulli GLM}}
  & \multicolumn{3}{c}{\textbf{SIR}} \\
\rowcolor{gray!10}
\textbf{Budget}
  & \textbf{10,000} & \textbf{20,000} & \textbf{30,000}
  & \textbf{10,000} & \textbf{20,000} & \textbf{30,000}
  & \textbf{10,000} & \textbf{20,000} & \textbf{30,000} \\
\hline

GATSBI
& 117.67 \tiny{$\pm$2.43}
& 164.90 \tiny{$\pm$57.10}
& 249.50 \tiny{$\pm$105.05}
& 118.62 \tiny{$\pm$0.18}
& 179.69 \tiny{$\pm$49.96}
& 253.48 \tiny{$\pm$113.38}
& 227.16 \tiny{$\pm$1.44}
& 241.24 \tiny{$\pm$7.58}
& 250.47 \tiny{$\pm$2.22} \\

NPE
& 7.04 \tiny{$\pm$0.21}
& 14.21 \tiny{$\pm$0.04}
& 20.95 \tiny{$\pm$0.37}
& 12.17 \tiny{$\pm$0.17}
& 24.96 \tiny{$\pm$0.64}
& 37.43 \tiny{$\pm$0.22}
& 20.10 \tiny{$\pm$5.45}
& 29.62 \tiny{$\pm$3.74}
& 44.92 \tiny{$\pm$9.24} \\

JANA
& 5.66 \tiny{$\pm$0.10}
& 9.48 \tiny{$\pm$0.12}
& 14.34 \tiny{$\pm$0.11}
& 7.14 \tiny{$\pm$0.05}
& 12.23 \tiny{$\pm$0.07}
& 19.00 \tiny{$\pm$0.19}
& 22.98 \tiny{$\pm$0.32}
& 39.17 \tiny{$\pm$2.48}
& 61.89 \tiny{$\pm$5.41} \\

APT
& 90.31 \tiny{$\pm$8.99}
& 186.17 \tiny{$\pm$13.79}
& 268.28 \tiny{$\pm$32.88}
& 101.39 \tiny{$\pm$7.89}
& 197.30 \tiny{$\pm$9.32}
& 267.38 \tiny{$\pm$21.93}
& 40.15 \tiny{$\pm$3.90}
& 73.74 \tiny{$\pm$6.25}
& 112.51 \tiny{$\pm$2.51} \\

Simformer
& 3.96 \tiny{$\pm$0.04}
& 7.36 \tiny{$\pm$0.11}
& 10.73 \tiny{$\pm$0.25}
& 17.87 \tiny{$\pm$0.18}
& 35.05 \tiny{$\pm$0.52}
& 52.81 \tiny{$\pm$0.91}
& 4.24 \tiny{$\pm$0.29}
& 7.67 \tiny{$\pm$0.38}
& 10.68 \tiny{$\pm$0.45} \\

\rowcolor{blue!5}
CP-VAE
& 2.23 \tiny{$\pm$0.81}
& 3.75 \tiny{$\pm$0.80}
& \textbf{4.99} \tiny{$\pm$0.83}
& \textbf{1.76} \tiny{$\pm$0.27}
& \textbf{3.64} \tiny{$\pm$0.14}
& \textbf{4.93} \tiny{$\pm$0.47}
& \textbf{2.77} \tiny{$\pm$0.40}
& \textbf{6.06} \tiny{$\pm$1.09}
& \textbf{8.68} \tiny{$\pm$0.83} \\

\rowcolor{blue!5}
UP-VAE
& \textbf{1.98} \tiny{$\pm$0.33}
& \textbf{3.27} \tiny{$\pm$0.58}
& 5.35 \tiny{$\pm$1.82}
& 2.10 \tiny{$\pm$0.50}
& 4.43 \tiny{$\pm$0.61}
& 6.57 \tiny{$\pm$1.04}
& 4.44 \tiny{$\pm$0.88}
& 6.71 \tiny{$\pm$1.97}
& 9.60 \tiny{$\pm$1.66} \\

\bottomrule
\end{tabular}
\end{adjustbox}
\end{subtable}

\vspace{0.5cm}
\begin{subtable}[t]{\textwidth}
\centering
\caption{SLCP Distractors, Bernoulli GLM Raw, Lotka Volterra.}
\begin{adjustbox}{max width=\textwidth}
\begin{tabular}{l|ccc|ccc|ccc}
\rowcolor{gray!10}
\textbf{Method}
  & \multicolumn{3}{c|}{\textbf{SLCP Distractors}}
  & \multicolumn{3}{c|}{\textbf{Bernoulli GLM Raw}}
  & \multicolumn{3}{c}{\textbf{Lotka Volterra}} \\
\rowcolor{gray!10}
\textbf{Budget}
  & \textbf{10,000} & \textbf{20,000} & \textbf{30,000}
  & \textbf{10,000} & \textbf{20,000} & \textbf{30,000}
  & \textbf{10,000} & \textbf{20,000} & \textbf{30,000} \\
\hline

GATSBI
& 74.37 \tiny{$\pm$2.59}
& 100.59 \tiny{$\pm$2.39}
& 102.30 \tiny{$\pm$0.51}
& 114.76 \tiny{$\pm$1.19}
& 115.22 \tiny{$\pm$1.43}
& 144.10 \tiny{$\pm$24.15}
& 78.91 \tiny{$\pm$6.32}
& 75.73 \tiny{$\pm$0.70}
& 74.99 \tiny{$\pm$0.20} \\

NPE
& 21.92 \tiny{$\pm$2.41}
& 43.92 \tiny{$\pm$2.58}
& 64.38 \tiny{$\pm$7.13}
& 25.39 \tiny{$\pm$2.09}
& 61.43 \tiny{$\pm$2.61}
& 85.94 \tiny{$\pm$6.55}
& 80.67 \tiny{$\pm$14.61}
& 160.92 \tiny{$\pm$42.82}
& 188.04 \tiny{$\pm$29.46} \\

JANA
& 8.49 \tiny{$\pm$0.06}
& 13.59 \tiny{$\pm$0.24}
& 20.56 \tiny{$\pm$0.12}
& 24.02 \tiny{$\pm$0.37}
& 44.21 \tiny{$\pm$0.23}
& 64.50 \tiny{$\pm$1.01}
& 7.59 \tiny{$\pm$0.02}
& 13.28 \tiny{$\pm$0.14}
& 18.55 \tiny{$\pm$0.08} \\

APT
& 26.30 \tiny{$\pm$0.36}
& 49.96 \tiny{$\pm$2.73}
& 74.31 \tiny{$\pm$2.54}
& 21.87 \tiny{$\pm$1.28}
& 40.00 \tiny{$\pm$1.37}
& 60.94 \tiny{$\pm$3.00}
& 45.52 \tiny{$\pm$4.85}
& 96.15 \tiny{$\pm$13.77}
& 144.33 \tiny{$\pm$18.33} \\

Simformer
& 15.98 \tiny{$\pm$1.22}
& 31.98 \tiny{$\pm$2.34}
& 47.85 \tiny{$\pm$1.86}
& 16.96 \tiny{$\pm$0.51}
& 32.46 \tiny{$\pm$1.89}
& 48.01 \tiny{$\pm$1.44}
& 5.0 \tiny{$\pm$0.34}
& 9.23 \tiny{$\pm$0.54}
& 13.64 \tiny{$\pm$0.21} \\

\rowcolor{blue!5}
CP-VAE
& \textbf{1.54} \tiny{$\pm$0.34}
& 3.62 \tiny{$\pm$0.76}
& 4.63 \tiny{$\pm$1.52}
& 6.04 \tiny{$\pm$0.99}
& 14.50 \tiny{$\pm$0.18}
& \textbf{10.36} \tiny{$\pm$0.20}
& 7.22 \tiny{$\pm$3.25}
& 14.61 \tiny{$\pm$1.92}
& 22.87 \tiny{$\pm$1.84} \\

\rowcolor{blue!5}
UP-VAE
& 1.89 \tiny{$\pm$0.15}
& \textbf{2.85} \tiny{$\pm$0.39}
& \textbf{4.25} \tiny{$\pm$0.52}
& \textbf{3.42} \tiny{$\pm$0.49}
& \textbf{8.64} \tiny{$\pm$2.80}
& 14.21 \tiny{$\pm$7.27}
& \textbf{2.89} \tiny{$\pm$0.52}
& \textbf{4.93} \tiny{$\pm$0.10}
& \textbf{7.16} \tiny{$\pm$1.26} \\

\bottomrule
\end{tabular}
\end{adjustbox}
\end{subtable}

\vspace{0.5cm}
\begin{subtable}{\textwidth}
\centering
\caption{Gaussian Mixture, Gaussian Linear, Gaussian Linear Uniform, and SLCP}
\renewcommand{\arraystretch}{1.1}
\setlength{\tabcolsep}{2pt}
\begin{adjustbox}{max width=\textwidth}
\begin{tabular}{l|ccc|ccc|ccc|ccc}
\rowcolor{gray!10}
\textbf{Method}
 & \multicolumn{3}{c|}{\textbf{Gaussian Mixture}}
 & \multicolumn{3}{c|}{\textbf{Gaussian Linear}}
 & \multicolumn{3}{c|}{\textbf{Gaussian Linear Uniform}}
 & \multicolumn{3}{c}{\textbf{SLCP}} \\

\rowcolor{gray!10}
\textbf{Budget}
 & \textbf{10,000} & \textbf{20,000} & \textbf{30,000}
 & \textbf{10,000} & \textbf{20,000} & \textbf{30,000}
 & \textbf{10,000} & \textbf{20,000} & \textbf{30,000}
 & \textbf{10,000} & \textbf{20,000} & \textbf{30,000} \\
\hline

GATSBI
& 114.07 \tiny{$\pm$2.11}
& 111.30 \tiny{$\pm$0.28}
& 137.46 \tiny{$\pm$19.16}
& 48.19 \tiny{$\pm$0.24}
& 48.70 \tiny{$\pm$0.84}
& 49.12 \tiny{$\pm$0.49}
& 43.27 \tiny{$\pm$0.36}
& 66.43 \tiny{$\pm$5.70}
& 69.10 \tiny{$\pm$0.15}
& 162.72 \tiny{$\pm$14.73}
& 196.19 \tiny{$\pm$119.42}
& 99.39 \tiny{$\pm$5.63} \\

NPE
& 19.56 \tiny{$\pm$2.47}
& 42.88 \tiny{$\pm$4.54}
& 64.03 \tiny{$\pm$6.51}
& 12.03 \tiny{$\pm$0.80}
& 24.69 \tiny{$\pm$1.63}
& 38.79 \tiny{$\pm$1.23}
& 16.55 \tiny{$\pm$0.85}
& 30.90 \tiny{$\pm$2.77}
& 49.42 \tiny{$\pm$1.49}
& 42.97 \tiny{$\pm$6.35}
& 97.70 \tiny{$\pm$6.66}
& 117.61 \tiny{$\pm$4.12} \\

JANA
& 7.19 \tiny{$\pm$0.03}
& 13.95 \tiny{$\pm$0.94}
& 17.14 \tiny{$\pm$0.10}
& \textbf{2.21} \tiny{$\pm$0.03}
& \textbf{3.68} \tiny{$\pm$0.05}
& \textbf{5.21} \tiny{$\pm$0.05}
& \textbf{1.82} \tiny{$\pm$0.02}
& \textbf{2.96} \tiny{$\pm$0.01}
& \textbf{4.10} \tiny{$\pm$0.03}
& 29.34 \tiny{$\pm$0.60}
& 53.76 \tiny{$\pm$1.05}
& 76.19 \tiny{$\pm$5.71} \\

APT
& 76.27 \tiny{$\pm$7.02}
& 167.88 \tiny{$\pm$15.99}
& 240.13 \tiny{$\pm$22.74}
& 70.47 \tiny{$\pm$0.88}
& 135.38 \tiny{$\pm$5.75}
& 203.27 \tiny{$\pm$5.97}
& 78.61 \tiny{$\pm$2.11}
& 158.57 \tiny{$\pm$9.32}
& 219.44 \tiny{$\pm$11.83}
& 93.40 \tiny{$\pm$1.26}
& 150.29 \tiny{$\pm$11.10}
& 216.30 \tiny{$\pm$21.38} \\

Simformer
& 3.84 \tiny{$\pm$0.04}
& 7.15 \tiny{$\pm$0.09}
& 10.48 \tiny{$\pm$0.21}
& 18.18 \tiny{$\pm$0.13}
& 35.46 \tiny{$\pm$0.17}
& 55.46 \tiny{$\pm$1.13}
& 17.89 \tiny{$\pm$0.23}
& 33.47 \tiny{$\pm$0.40}
& 51.24 \tiny{$\pm$0.89}
& 4.23 \tiny{$\pm$0.01}
& \textbf{7.67} \tiny{$\pm$0.13}
& \textbf{11.32} \tiny{$\pm$0.40} \\

\rowcolor{blue!5}
CP-VAE
& \textbf{2.37} \tiny{$\pm$0.81}
& \textbf{4.67} \tiny{$\pm$1.10}
& \textbf{5.19} \tiny{$\pm$0.91}
& 3.11 \tiny{$\pm$0.58}
& 5.25 \tiny{$\pm$0.63}
& 6.49 \tiny{$\pm$1.38}
& 3.36 \tiny{$\pm$0.59}
& 7.48 \tiny{$\pm$1.65}
& 8.45 \tiny{$\pm$2.47}
& \textbf{3.85} \tiny{$\pm$0.62}
& 11.38 \tiny{$\pm$1.97}
& 19.53 \tiny{$\pm$2.99} \\

\rowcolor{blue!5}
UP-VAE
& 5.61 \tiny{$\pm$1.20}
& 6.30 \tiny{$\pm$1.29}
& 12.96 \tiny{$\pm$1.41}
& 5.75 \tiny{$\pm$1.16}
& 9.15 \tiny{$\pm$1.03}
& 9.16 \tiny{$\pm$1.74}
& 7.41 \tiny{$\pm$1.85}
& 14.08 \tiny{$\pm$1.55}
& 18.64 \tiny{$\pm$6.57}
& 5.49 \tiny{$\pm$0.65}
& 12.59 \tiny{$\pm$1.17}
& 22.69 \tiny{$\pm$3.23} \\

\bottomrule
\end{tabular}
\end{adjustbox}
\end{subtable}
\end{table}
\section{Discussion}
\label{sec:discussion}

The results are now evaluated with respect to the following important considerations in SBI.

    {\bf Posterior estimation accuracy:} Flow-based (NPE, APT, JANA) and diffusion-based methods (Simformer) typically yield higher accuracy, with Simformer often performing best overall. This is expected, as flow and diffusion frameworks tend to model complex posterior distributions more precisely than variational or GAN-based methods, which can suffer from oversimplified approximations, bound-related errors, and adversarial instability.
    
    {\bf Computational efficiency:} Adversarial training presents well known challenges, which reflect in relatively high training times for GATSBI. The sequential nature of APT adds computational expense over NPE and CP/UP-VAE. Autoregressive models can be computationally expensive to train and evaluate as estimating each parameter depends on previously estimated parameters -- however, JANA is able to closely match NPE's training times for chosen hyperparameters. Simformer, leveraging transformers, has high training time due to the complexity of transformer training and suffers from slow inference speed due to its iterative denoising process. CP-VAE and UP-VAE have simple yet flexible architectures enabling efficient and fast training (Table \ref{tab:train_times_all}).

    {\bf Robustness/stability:} GATSBI suffers from the instabilities associated with the adversarial setting (e.g., high MMD std. dev. in Table \ref{tab:mmd_global}). JANA, APT, Simformer proved to be very robust, with low std. dev. across test problems, followed by CP/UP-VAE and NPE.

    {\bf C2ST v/s MMD:} While the C2ST metric is a popular metric for validating inference models in SBI, it may be difficult to interpret in high dimensions, where one poorly estimated dimension can skew the score and results depend on classifier parameters. In contrast, MMD compares empirical distributions directly without relying on a separate classifier, providing a more robust measure in complex, high-dimensional posterior evaluations.

    {\bf Flexibility/approximating complex posteriors:} Simformer is highly effective overall, leveraging transformers and cross-attention conditioning mechanisms to learn discriminative patterns effectively. APT holds an advantage over NPE due to its iterative refinement. JANA is particularly suited for posteriors with dependencies between parameters. However, the flexibility of normalizing flow based approaches (NPE, APT, JANA) is hindered by the invertibility requirement. GATSBI is suitable for complex, high-dimensional problems where other approaches may struggle and learning the posterior \emph{implicitly} becomes important. Both CP-VAE and UP-VAE are highly flexible and adaptable with respect to the architecture.

{\bf CP-VAE v/s UP-VAE:} CP-VAE’s conditional prior \( p(\mathbf{z} \mid \mathbf{y}) \) adapts to observed data, allowing flexibility and control for improved posterior accuracy. This is of particular interest in high-dimensional problems. UP-VAE’s simple Gaussian prior \( p(\mathbf{z}) \) allows for efficient latent space sampling, easier regularization, and a consistent latent space, leading to more stable training, making it efficient and effective for simpler problems.

The proposed CP-VAE and UP-VAE models complement existing approaches by virtue of approximating the posterior distribution in a computation-efficient manner (as evidenced in Table \ref{tab:train_times_all}). This makes the models well-suited for challenging problems where computational efficiency is of key consideration (e.g., large number of parameters to infer, rapid model exploration, etc.). While UP-VAE uses a simpler prior, its data decoder is computationally more expensive than the conditional prior in CP-VAE, which reflects in faster training times for CP-VAE for most problems (Table \ref{tab:train_times_all}).
While we note that the training times in Table \ref{tab:train_times_all}) can be affected by several factors including implementation and hyperparameter optimization, the comparison provides an estimate of relative computational complexity on the same hardware for reference implementations (taken from the original papers). 

\section{Conclusion}
\label{Conclusion}
This paper introduces two latent variable models, CP-VAE and UP-VAE, for amortized inference in likelihood-free simulation-based inference, with a view on efficiency, interpretability and robustness. The study compares four generative model classes—VAEs, GANs, flow-based methods, and diffusion models—specifically tailored for simulation-based inference, evaluating them on a standardized benchmark for accuracy and training speed. Both CP-VAE and UP-VAE are able to approximate complex posterior distributions while being highly computational efficient and interpretable. 

Future enhancements could involve integrating more sophisticated architectures (e.g., Convolutional Neural Networks) to better capture spatial and temporal dependencies within the data. Additionally, incorporating normalizing flows can provide more flexible posterior approximations, enhancing the model's ability to represent complex distributions. Exploring a variety of prior distributions, including truncated Gaussian priors, may further improve the model's capacity to approximate intricate posterior distributions.

\section{Acknowledgements}
The computations/data handling were enabled by the Berzelius resource
provided by the Knut and Alice Wallenberg Foundation at
the National Supercomputer Centre and by the National
Academic Infrastructure for Supercomputing in Sweden
(NAISS) at Chalmers e-Commons at Chalmers, and Uppsala Multidisciplinary Center for Advanced Computational
Science (UPPMAX) at Uppsala University. Andreas Hellander and Prashant Singh acknowledge
support from the Swedish Research Council through grant
agreement nos. 2023-05167 and 2023-05593 respectively.


\bibliography{ref}

\begin{thebibliography}{26}
\providecommand{\natexlab}[1]{#1}
\providecommand{\url}[1]{\texttt{#1}}
\expandafter\ifx\csname urlstyle\endcsname\relax
  \providecommand{\doi}[1]{doi: #1}\else
  \providecommand{\doi}{doi: \begingroup \urlstyle{rm}\Url}\fi

\bibitem[Arjovsky and Bottou(2022)]{Arjovsky17a}
Martin Arjovsky and Leon Bottou.
\newblock Towards principled methods for training generative adversarial networks, 2022.

\bibitem[Arjovsky et~al.(2017)Arjovsky, Chintala, and Bottou]{Arjovsky17b}
Martin Arjovsky, Soumith Chintala, and L{\'e}on Bottou.
\newblock Wasserstein generative adversarial networks.
\newblock In \emph{International conference on machine learning}, pages 214--223. PMLR, 2017.

\bibitem[Cranmer et~al.(2020)Cranmer, Brehmer, and Louppe]{Cranmer20}
Kyle Cranmer, Johann Brehmer, and Gilles Louppe.
\newblock The frontier of simulation-based inference.
\newblock \emph{Proceedings of the National Academy of Sciences}, 117\penalty0 (48):\penalty0 30055--30062, 2020.

\bibitem[Friedman(2003)]{Friedman:2003id}
Jerome~H Friedman.
\newblock On multivariate goodness-of-fit and two-sample testing.
\newblock \emph{Statistical Problems in Particle Physics, Astrophysics, and Cosmology}, 1:\penalty0 311, 2003.

\bibitem[Gl{\"o}ckler et~al.(2022)Gl{\"o}ckler, Deistler, and Macke]{glöckler2022variational}
Manuel Gl{\"o}ckler, Michael Deistler, and Jakob~H. Macke.
\newblock Variational methods for simulation-based inference.
\newblock In \emph{International Conference on Learning Representations}, 2022.
\newblock URL \url{https://openreview.net/forum?id=kZ0UYdhqkNY}.

\bibitem[Gloeckler et~al.(2024)Gloeckler, Deistler, Weilbach, Wood, and Macke]{gloecklerall}
Manuel Gloeckler, Michael Deistler, Christian Weilbach, Frank Wood, and Jakob~H. Macke.
\newblock All-in-one simulation-based inference.
\newblock In \emph{Proceedings of the 41st International Conference on Machine Learning}, ICML'24. JMLR.org, 2024.

\bibitem[Greenberg et~al.(2019)Greenberg, Nonnenmacher, and Macke]{Greenberg19}
David Greenberg, Marcel Nonnenmacher, and Jakob Macke.
\newblock Automatic posterior transformation for likelihood-free inference.
\newblock In \emph{International Conference on Machine Learning}, pages 2404--2414. PMLR, 2019.

\bibitem[Hermans et~al.(2020)Hermans, Begy, and Louppe]{hermans2020likelihoodfree}
Joeri Hermans, Volodimir Begy, and Gilles Louppe.
\newblock Likelihood-free mcmc with amortized approximate ratio estimators.
\newblock In \emph{International conference on machine learning}, pages 4239--4248. PMLR, 2020.

\bibitem[Ivanov et~al.(2019)Ivanov, Figurnov, and Vetrov]{ivanov2018variational}
Oleg Ivanov, Michael Figurnov, and Dmitry Vetrov.
\newblock Variational autoencoder with arbitrary conditioning.
\newblock In \emph{International Conference on Learning Representations}, 2019.
\newblock URL \url{https://openreview.net/forum?id=SyxtJh0qYm}.

\bibitem[Jordan et~al.(1999)Jordan, Ghahramani, Jaakkola, and Saul]{jordan1999introduction}
Michael~I Jordan, Zoubin Ghahramani, Tommi~S Jaakkola, and Lawrence~K Saul.
\newblock An introduction to variational methods for graphical models.
\newblock \emph{Machine learning}, 37\penalty0 (2):\penalty0 183--233, 1999.
\newblock \doi{10.1023/A:1007665907178}.

\bibitem[Kingma and Welling(2014)]{kingma2022autoencodingvariationalbayes}
Diederik~P. Kingma and Max Welling.
\newblock Auto-encoding variational bayes.
\newblock In \emph{2nd International Conference on Learning Representations, {ICLR} 2014}, 2014.
\newblock URL \url{http://arxiv.org/abs/1312.6114v10}.

\bibitem[Loshchilov and Hutter(2019)]{loshchilov2019decoupled}
Ilya Loshchilov and Frank Hutter.
\newblock Decoupled weight decay regularization, 2019.

\bibitem[Lueckmann et~al.(2017)Lueckmann, Goncalves, Bassetto, Öcal, Nonnenmacher, and Macke]{NIPS2017_addfa9b7}
Jan-Matthis Lueckmann, Pedro~J Goncalves, Giacomo Bassetto, Kaan Öcal, Marcel Nonnenmacher, and Jakob~H Macke.
\newblock Flexible statistical inference for mechanistic models of neural dynamics.
\newblock In \emph{Advances in Neural Information Processing Systems}, volume~30. Curran Associates, Inc., 2017.
\newblock URL \url{https://proceedings.neurips.cc/paper_files/paper/2017/file/addfa9b7e234254d26e9c7f2af1005cb-Paper.pdf}.

\bibitem[Lueckmann et~al.(2021)Lueckmann, Boelts, Greenberg, Goncalves, and Macke]{lueckmann2021benchmarking}
Jan-Matthis Lueckmann, Jan Boelts, David Greenberg, Pedro Goncalves, and Jakob Macke.
\newblock Benchmarking simulation-based inference.
\newblock In \emph{Proceedings of The 24th International Conference on Artificial Intelligence and Statistics}, volume 130 of \emph{Proceedings of Machine Learning Research}, pages 343--351. PMLR, 13--15 Apr 2021.

\bibitem[Papamakarios and Murray(2016)]{papamakarios2016fast}
George Papamakarios and Iain Murray.
\newblock Fast $\varepsilon$-free inference of simulation models with bayesian conditional density estimation.
\newblock \emph{Advances in neural information processing systems}, 29, 2016.

\bibitem[Radev et~al.(2023{\natexlab{a}})Radev, Schmitt, Pratz, Picchini, K{\"o}the, and B{\"u}rkner]{Radev23}
Stefan~T Radev, Marvin Schmitt, Valentin Pratz, Umberto Picchini, Ullrich K{\"o}the, and Paul-Christian B{\"u}rkner.
\newblock Jana: Jointly amortized neural approximation of complex bayesian models.
\newblock In \emph{Uncertainty in Artificial Intelligence}, pages 1695--1706. PMLR, 2023{\natexlab{a}}.

\bibitem[Radev et~al.(2023{\natexlab{b}})Radev, Schmitt, Schumacher, Elsemüller, Pratz, Schälte, Köthe, and Bürkner]{bayesflow_2023_software}
Stefan~T. Radev, Marvin Schmitt, Lukas Schumacher, Lasse Elsemüller, Valentin Pratz, Yannik Schälte, Ullrich Köthe, and Paul-Christian Bürkner.
\newblock {BayesFlow}: Amortized {B}ayesian workflows with neural networks.
\newblock \emph{Journal of Open Source Software}, 8\penalty0 (89):\penalty0 5702, 2023{\natexlab{b}}.

\bibitem[Ramesh et~al.(2022)Ramesh, Lueckmann, Boelts, Tejero-Cantero, Greenberg, Goncalves, and Macke]{Ramesh22}
Poornima Ramesh, Jan-Matthis Lueckmann, Jan Boelts, {\'A}lvaro Tejero-Cantero, David~S. Greenberg, Pedro~J. Goncalves, and Jakob~H. Macke.
\newblock {GATSBI}: Generative adversarial training for simulation-based inference.
\newblock In \emph{International Conference on Learning Representations}, 2022.
\newblock URL \url{https://openreview.net/forum?id=kR1hC6j48Tp}.

\bibitem[Rezende and Mohamed(2015)]{flow}
Danilo Rezende and Shakir Mohamed.
\newblock Variational inference with normalizing flows.
\newblock In \emph{International conference on machine learning}, pages 1530--1538. PMLR, 2015.

\bibitem[Sheth and Kahou(2024)]{10.5555/3666122.3667294}
Ivaxi Sheth and Samira~Ebrahimi Kahou.
\newblock Auxiliary losses for learning generalizable concept-based models.
\newblock In \emph{Proceedings of the 37th International Conference on Neural Information Processing Systems}, NeurIPS '23, Red Hook, NY, USA, 2024. Curran Associates Inc.

\bibitem[Sisson et~al.(2018)Sisson, Fan, and Beaumont]{sisson2018handbook}
Scott~A Sisson, Yanan Fan, and Mark Beaumont.
\newblock \emph{Handbook of approximate Bayesian computation}.
\newblock CRC press, 2018.

\bibitem[Sohn et~al.(2015)Sohn, Lee, and Yan]{NIPS2015_8d55a249}
Kihyuk Sohn, Honglak Lee, and Xinchen Yan.
\newblock Learning structured output representation using deep conditional generative models.
\newblock \emph{Advances in neural information processing systems}, 28, 2015.

\bibitem[Tejero-Cantero et~al.(2020)Tejero-Cantero, Boelts, Deistler, Lueckmann, Durkan, Gonçalves, Greenberg, and Macke]{tejero-cantero2020sbi}
Alvaro Tejero-Cantero, Jan Boelts, Michael Deistler, Jan-Matthis Lueckmann, Conor Durkan, Pedro~J. Gonçalves, David~S. Greenberg, and Jakob~H. Macke.
\newblock sbi: A toolkit for simulation-based inference.
\newblock \emph{Journal of Open Source Software}, 5\penalty0 (52):\penalty0 2505, 2020.
\newblock \doi{10.21105/joss.02505}.
\newblock URL \url{https://doi.org/10.21105/joss.02505}.

\bibitem[Ward et~al.(2022)Ward, Cannon, Beaumont, Fasiolo, and Schmon]{Ward22}
Daniel Ward, Patrick Cannon, Mark Beaumont, Matteo Fasiolo, and Sebastian Schmon.
\newblock Robust neural posterior estimation and statistical model criticism.
\newblock \emph{Advances in Neural Information Processing Systems}, 35:\penalty0 33845--33859, 2022.

\bibitem[Wiqvist et~al.(2021)Wiqvist, Frellsen, and Picchini]{wiqvist2021sequential}
Samuel Wiqvist, Jes Frellsen, and Umberto Picchini.
\newblock Sequential neural posterior and likelihood approximation, 2021.

\bibitem[Zammit-Mangion et~al.(2024)Zammit-Mangion, Sainsbury-Dale, and Huser]{zammit2024neural}
Andrew Zammit-Mangion, Matthew Sainsbury-Dale, and Rapha{\"e}l Huser.
\newblock Neural methods for amortised parameter inference.
\newblock \emph{arXiv preprint arXiv:2404.12484}, 2024.

\end{thebibliography}

\newpage

\onecolumn

\title{Supplementary Material}
\maketitle


\appendix
\section{Hodgkin-Huxley Model}
\begin{figure}[h]
    \centering
    \begin{subfigure}[t]{0.48\textwidth}
        \centering
        \includegraphics[width=\textwidth]{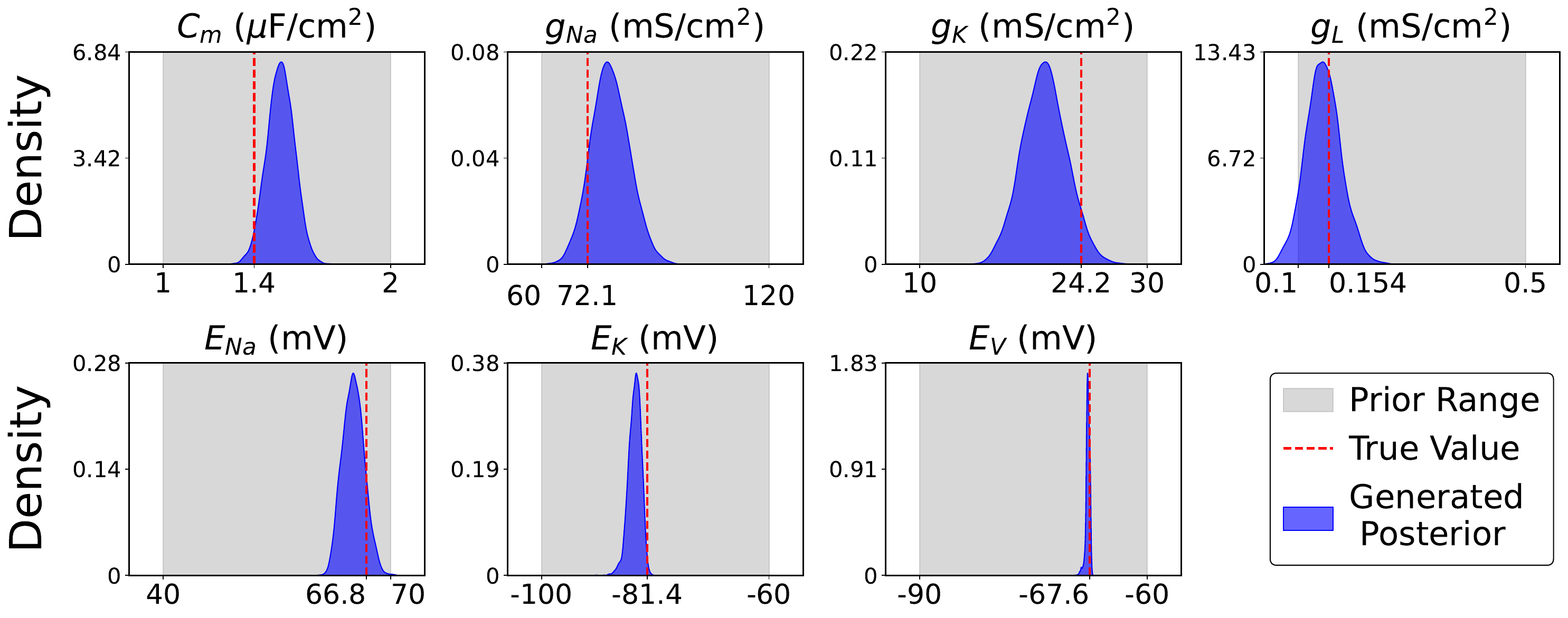}
        \caption{CP-VAE}
        \label{fig:hh_v1}
    \end{subfigure}
    \hfill
    \vspace{0.5cm}
    \begin{subfigure}[t]{0.48\textwidth}
        \centering
        \includegraphics[width=\textwidth]{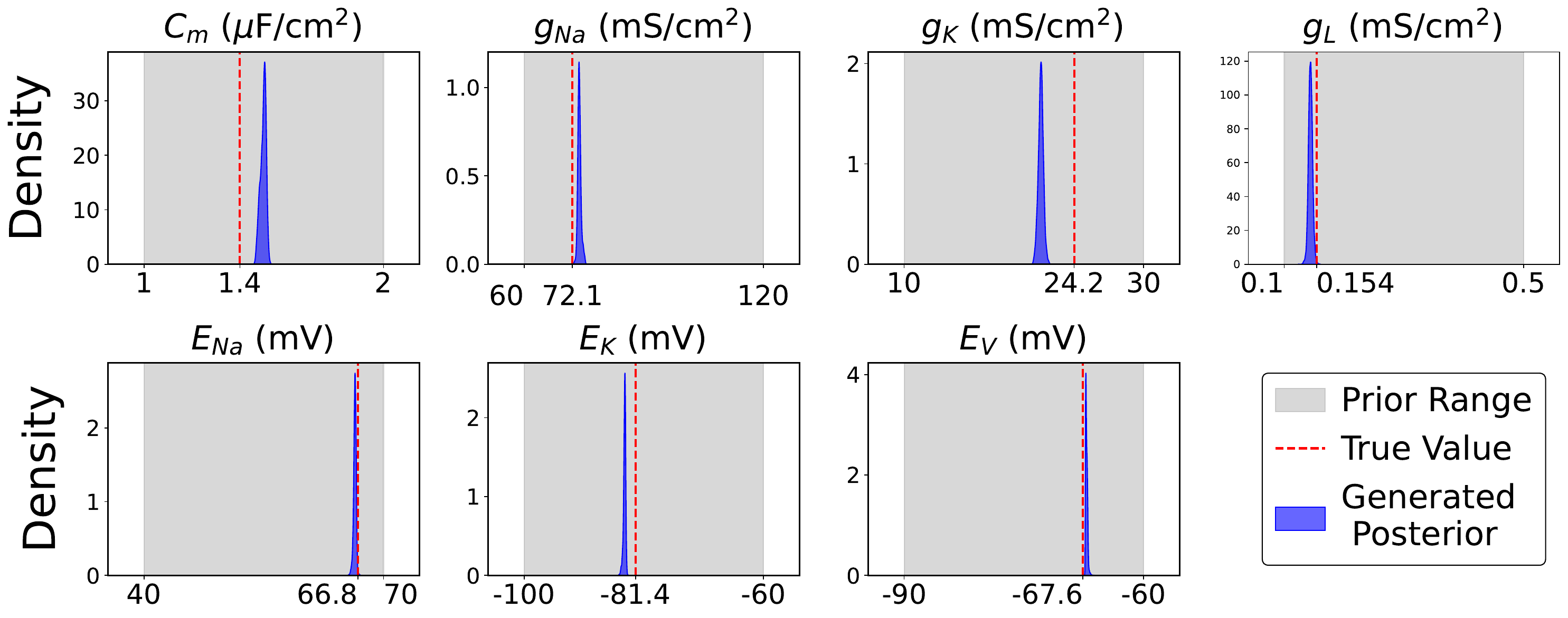}
        \caption{UP-VAE}
        \label{fig:hh_v2}
    \end{subfigure}
    \caption{HH Model: Estimated posterior (simulation budget: 10,000). (a) CP-VAE and (b) UP-VAE.}
    \label{fig:hh_posterior}
\end{figure}

\begin{figure}[h]
    \centering
    \begin{subfigure}[t]{0.4\textwidth}
        \centering
        \includegraphics[width=\textwidth]{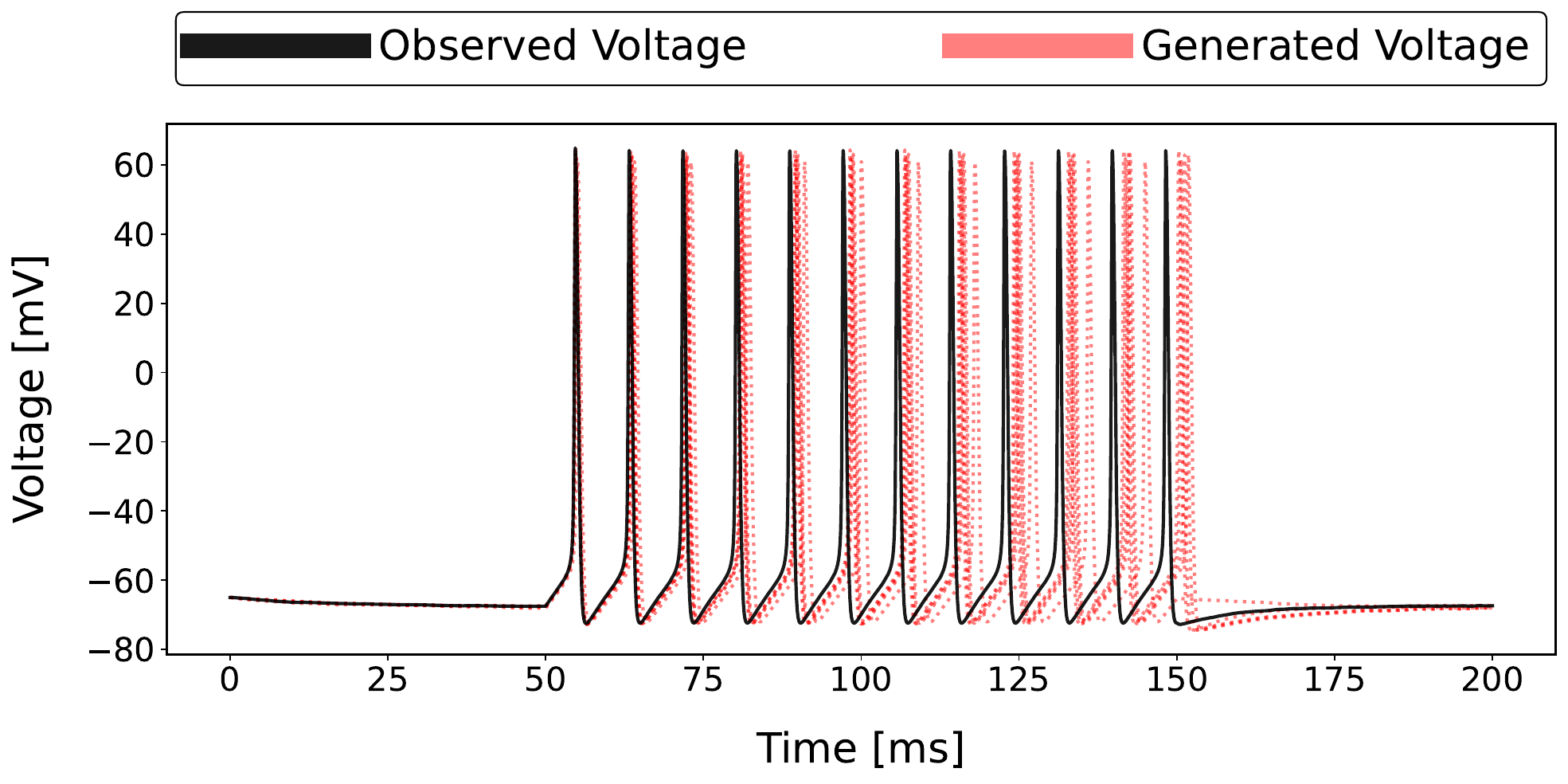}
        \caption{CP-VAE}
        \label{fig:hh_v1}
    \end{subfigure}
    \hspace{1mm}
    \begin{subfigure}[t]{0.4\textwidth}
        \centering
        \includegraphics[width=\textwidth]{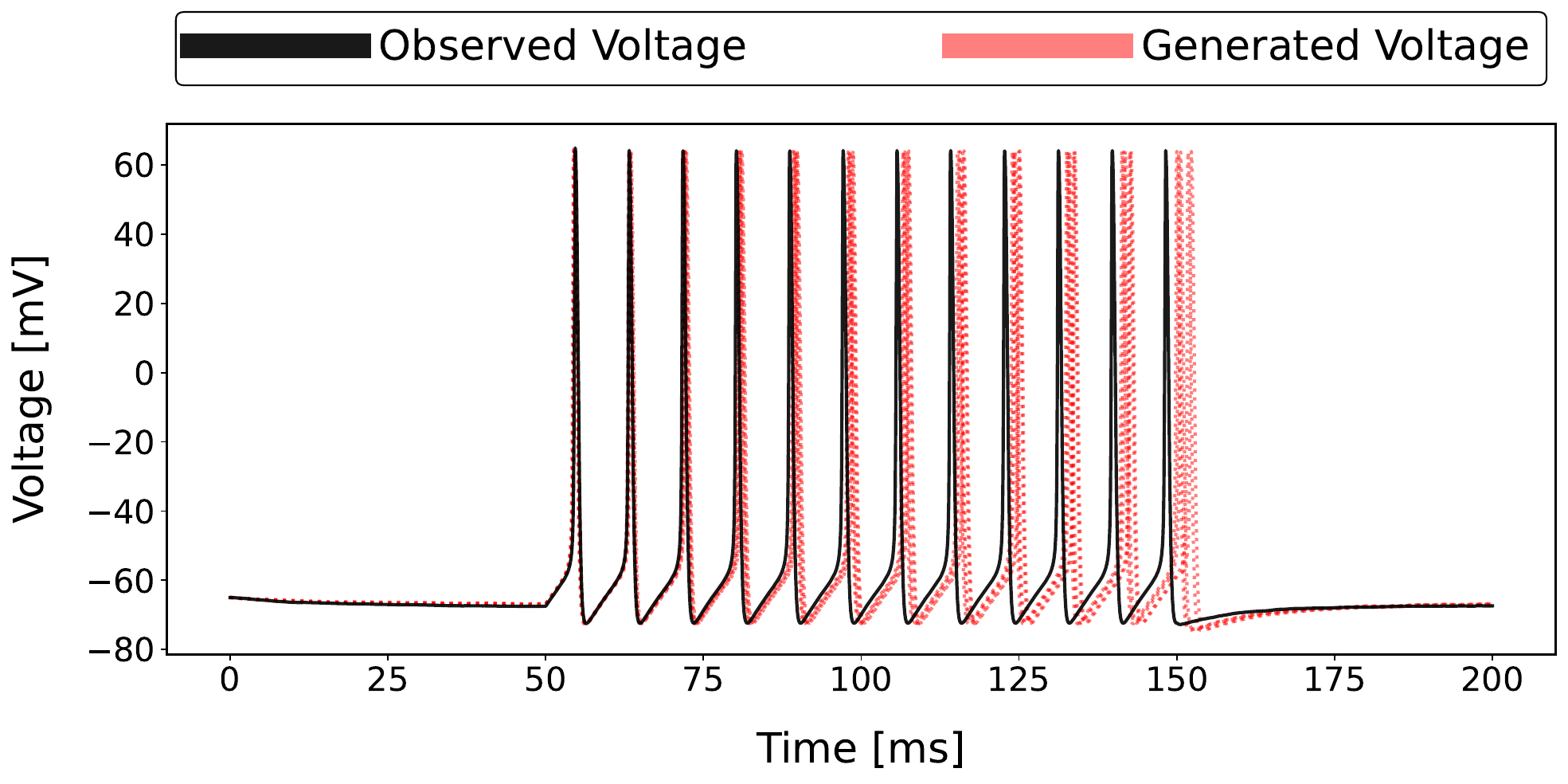}
        \caption{UP-VAE}
        \label{fig:hh_v2}
    \end{subfigure}
    \caption{HH Model: Posterior predictive checks (simulation budget: 10,000). (a) CP-VAE and (b) UP-VAE.}
    \label{fig:hh_voltage}
\end{figure}

We evaluate our variational methods on the Hodgkin-Huxley (HH) model \citep{gloecklerall}, which simulates neuronal action potentials by modeling membrane voltage under sodium, potassium, and leak currents. The parameter vector \(\boldsymbol{\theta} = (C_m, g_{\text{Na}}, g_{\text{K}}, g_{\text{L}}, E_{\text{Na}}, E_{\text{K}}, E_{\text{L}})\) is sampled from a uniform prior, and the system is run for 200\,ms with a brief current injection. Since the true posterior is unknown, we evaluate our parameter inference performance by generating 10{,}000 samples from each model for a specific parameter configuration \((1.437, 72.177, 24.209, 0.154, 66.87, -81.435, -67.606)\) and by conducting a posterior predictive check—randomly selecting five posterior draws to simulate new voltage traces. As shown in Figure~\ref{fig:hh_posterior}, both models provide reasonable posterior estimates, with the UP-VAE generally exhibiting lower variance, while Figure~\ref{fig:hh_voltage} indicates that the generated voltage trajectories closely match the reference voltage signals. The training time for CP-VAE is \(4.7 \pm 1.17\) minutes, while UP-VAE trains in \(5.03 \pm 0.04\) minutes, averaged over 5 independent runs with a budget of 10{,}000.

\section{Hyperparameters}
For JANA and Simformer, we use hyperparameters published in their original works \citep{Radev23,gloecklerall}. The NPE approach adopts normalizing spline flows (NSF) from the \texttt{sbi} toolbox, APT follows the same NSF setup \citep{tejero-cantero2020sbi} but distributes each budget across multiple rounds, dividing total simulations by the number of rounds. For GATSBI, hyperparameters for available tasks were taken directly from \citep{Ramesh22}, while for other test problems, we adapt them according to each problem’s structural and computational requirements. The configurations for UP-VAE and CP-VAE are described below:

\subsection{\textbf{CP-VAE}}
In this configuration, common parameters are applied uniformly across all problems, with specific architectures detailed in Table~\ref{tab:cpvae_hyper}. Each network — \(q_{\boldsymbol{\phi}}(\boldsymbol{z} \mid \boldsymbol{y}, \boldsymbol{\theta})\), \(p(\boldsymbol{z} \mid \boldsymbol{y})\), and \(p(\boldsymbol{\theta} \mid \boldsymbol{y}, \boldsymbol{z})\) — is designed to approximate a multivariate distribution, outputting both a mean and variance vector. Specifically, the last layer in \(q_{\boldsymbol{\phi}}(\boldsymbol{z} \mid \boldsymbol{y}, \boldsymbol{\theta})\) and \(p(\boldsymbol{z} \mid \boldsymbol{y})\) outputs a vector of size \(2 \times \text{latent\_dim}\), while the last layer in \(p(\boldsymbol{\theta} \mid \boldsymbol{y}, \boldsymbol{z})\) outputs a vector of size \(2 \times \text{theta\_dim}\). Leaky ReLU activation functions with a negative slope of 0.1 are used throughout, with a learning rate of \(5 \times 10^{-4}\), except for the Lotka-Volterra model, which uses \(10^{-4}\), and the AdamW optimizer \citep{loshchilov2019decoupled}. Models are trained with a batch size of 32 (except SLCP, which uses 16, and Lotka-Volterra, which uses 128) for 1,000 epochs, with early stopping triggered if validation performance does not improve for 20 consecutive epochs. Both parameters and the data are normalized using standard scaling, and gradient clipping is set to 3.0. The latent dimension is equal to the parameter dimension, except for the Gaussian Linear and GLM raw models, which use 5, Lotka-Volterra uses 8, SLCP uses 4, SLCP with Distractors uses 8, and Gaussian Linear uses 5. Kaiming normal initialization is used for weights. For each simulation budget, 90\% of the data is used for training, with the remaining 10\% reserved for validation.

\begin{table}[h]
\centering
\scriptsize 
\setlength{\tabcolsep}{3pt} 
\renewcommand{\arraystretch}{0.85} 
\caption{Architectural details of the CP-VAE model.}
\label{tab:cpvae_hyper}
\begin{tabular}{>{\columncolor{gray!15}}l c c c}
\toprule
\textbf{Problem} & \(\boldsymbol{q_{\phi}(z \mid y,\theta)}\) & \(\boldsymbol{p(z \mid y)}\) & \(\boldsymbol{p(\theta \mid y, z)}\) \\
\midrule
Gaussian Linear          & (64,64,64)           & (64,64)       & (64,64,64)          \\
Gaussian Lin. Uniform    & (64,64,64)           & (64,64)       & (64,64,64)          \\
SLCP                     & (200,200,200)        & (64,64)       & (200,200)           \\
SLCP + Distractors       & (200,200,200)        & (64,64)       & (200,200,200)       \\
Bernoulli GLM            & (128,128,128)        & (128,128)     & (128,128,128)       \\
Bernoulli GLM Raw        & (256,256,256,256)    & (128,128)     & (256,256,256,256)   \\
Gaussian Mixture         & (64,64,64)           & (64,64)       & (64,64,64)          \\
Two Moons                & (128,128,128)        & (128,128)     & (128,128,128)       \\
SIR                      & (100,100)            & (100,100)     & (100,100,100)       \\
Lotka-Volterra           & (128,128,128,128)    & (64,64)       & (128,128,128,128)   \\
HH Model                 & (128,128,128,128)    & (128,128,128,128) & (128,128,128,128) \\
\bottomrule
\end{tabular}
\end{table}

\subsection{\textbf{UP-VAE}}  
Building on CP-VAE, UP-VAE retains core training hyperparameters while introducing key refinements. The architectures of \( q_{\phi}(\boldsymbol{z} \mid \boldsymbol{y}, \boldsymbol{\theta}) \), \( p(\boldsymbol{\theta} \mid \boldsymbol{y}, \boldsymbol{z}) \), and \( p(\boldsymbol{y} \mid \boldsymbol{z}) \) are outlined in Table~\ref{tab:upvae_hyper}. The loss function weights are adjusted to prioritize the theta decoder and latent space constraint, with data reconstruction set to 0.2, theta loss to 0.4, and KL divergence to 0.4.

\begin{table}[h]
\centering
\scriptsize 
\setlength{\tabcolsep}{3pt} 
\renewcommand{\arraystretch}{0.85} 
\caption{Architectural details of the UP-VAE Model.}
\label{tab:upvae_hyper}
\begin{tabular}{>{\columncolor{gray!15}}l c c c}
\toprule
\textbf{Problem} & \(\boldsymbol{q_{\phi}(z \mid y,\theta)}\) & \(\boldsymbol{p(\theta \!\mid\! y,z)}\) & \(\boldsymbol{p(y\!\mid\!z)}\) \\
\midrule
Gauss. Linear        & (64,64,64)      & (64,64,64)      & (64,64,64)     \\
Gauss. Lin. Uniform  & (64,64,64)      & (64,64,64)      & (64,64,64)     \\
SLCP                 & (200,200,200)   & (200,200,200)   & (200,200,200)  \\
SLCP + Distractors   & (200,100,50,25) & (200,100,50,25) & (128,128,128)  \\
Bernoulli GLM        & (128,128,128)   & (128,128,128)   & (128,128,128)  \\
Bernoulli GLM Raw    & (128,128,128,128) & (128,128,128,128) & (128,128,128,128) \\
Gaussian Mixture     & (64,64,64)      & (64,64,64)      & (64,64,64)     \\
Two Moons            & (256,128,64)    & (256,128,64)    & (256,128,64)   \\
SIR                  & (100,100,100)   & (100,100,100)   & (100,100,100)  \\
Lotka-Volterra       & (128,128,128,128) & (128,128,128)    & (128,128,128)  \\
HH Model             & (256,128,64)    & (256,128,64)    & (256,128,64)   \\
\bottomrule
\end{tabular}
\end{table}


\newpage
\section{Additional Plots}
\begin{figure}[h]
    \centering
    \begin{subfigure}[t]{0.5\textwidth}
        \centering
        \includegraphics[width=\textwidth]{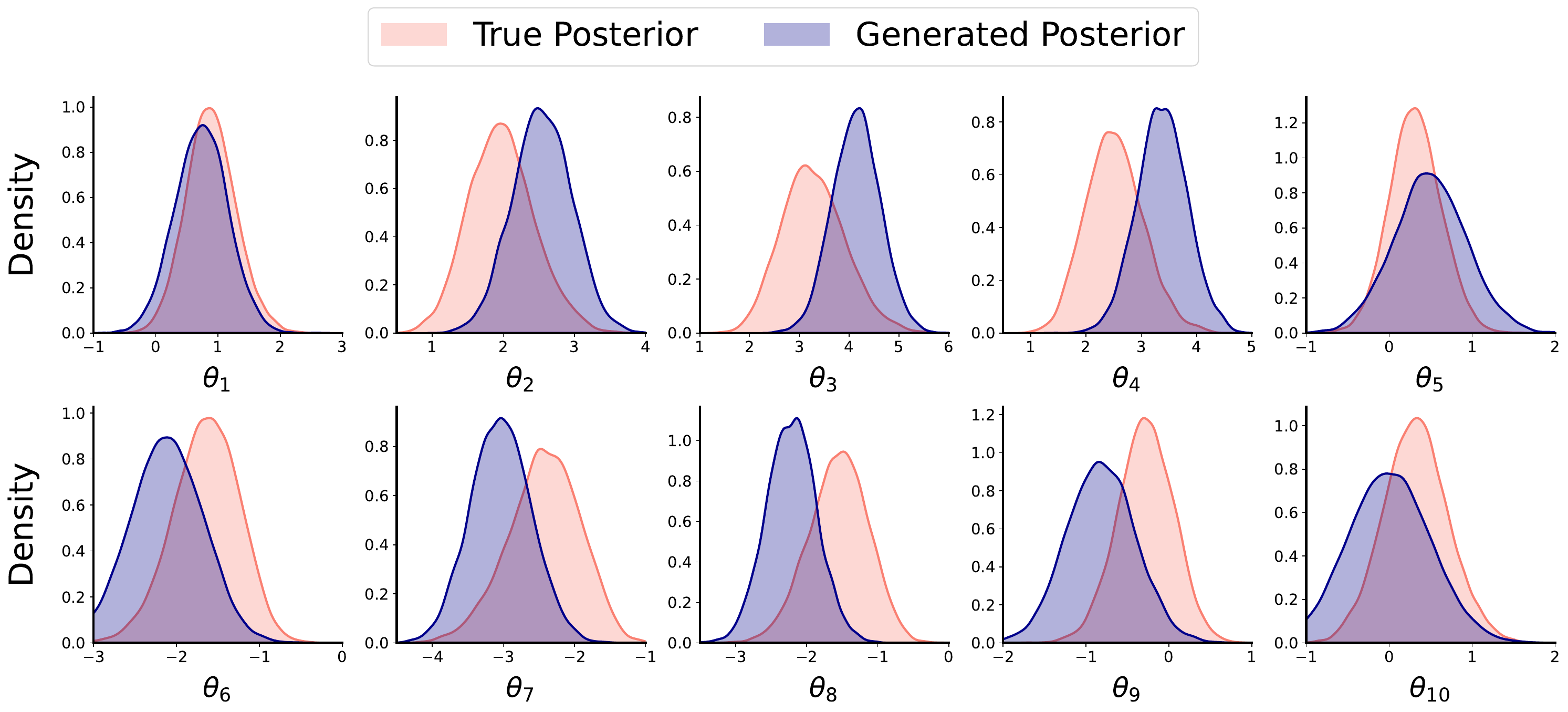}
        \caption{CP-VAE}
        \label{fig:glm_raw_v1}
    \end{subfigure}
    \hfill
    \begin{subfigure}[t]{0.5\textwidth}
        \centering
        \includegraphics[width=\textwidth]{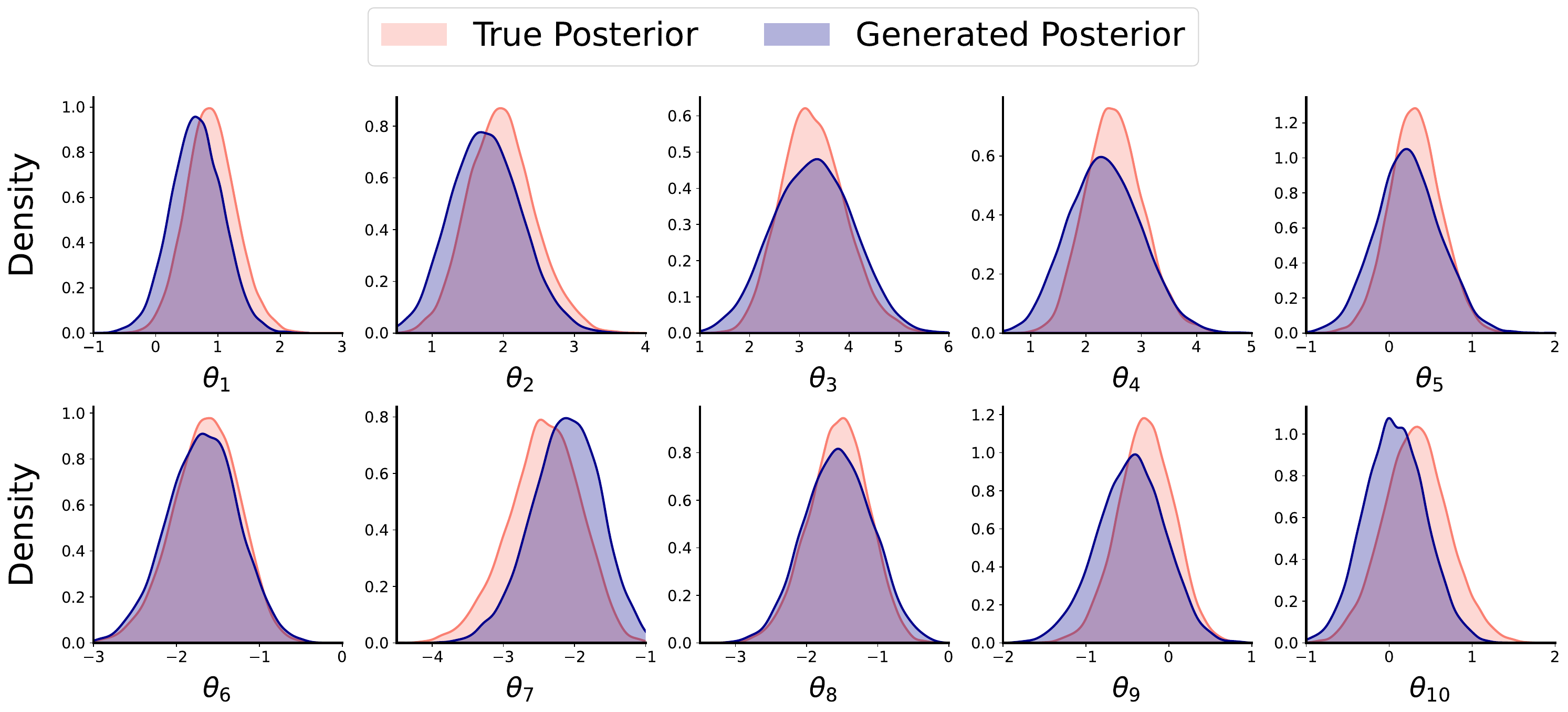}
        \caption{UP-VAE}
        \label{fig:glm_raw_v2}
    \end{subfigure}
    \caption{Bernoulli GLM Raw: Estimated posterior (simulation budget: 30,000). (a) CP-VAE and (b) UP-VAE.}
    \label{fig:glm_raw}
\end{figure}

\begin{figure}[h]
    \centering
    \begin{subfigure}[t]{0.4\textwidth}
        \centering
        \includegraphics[width=\textwidth]{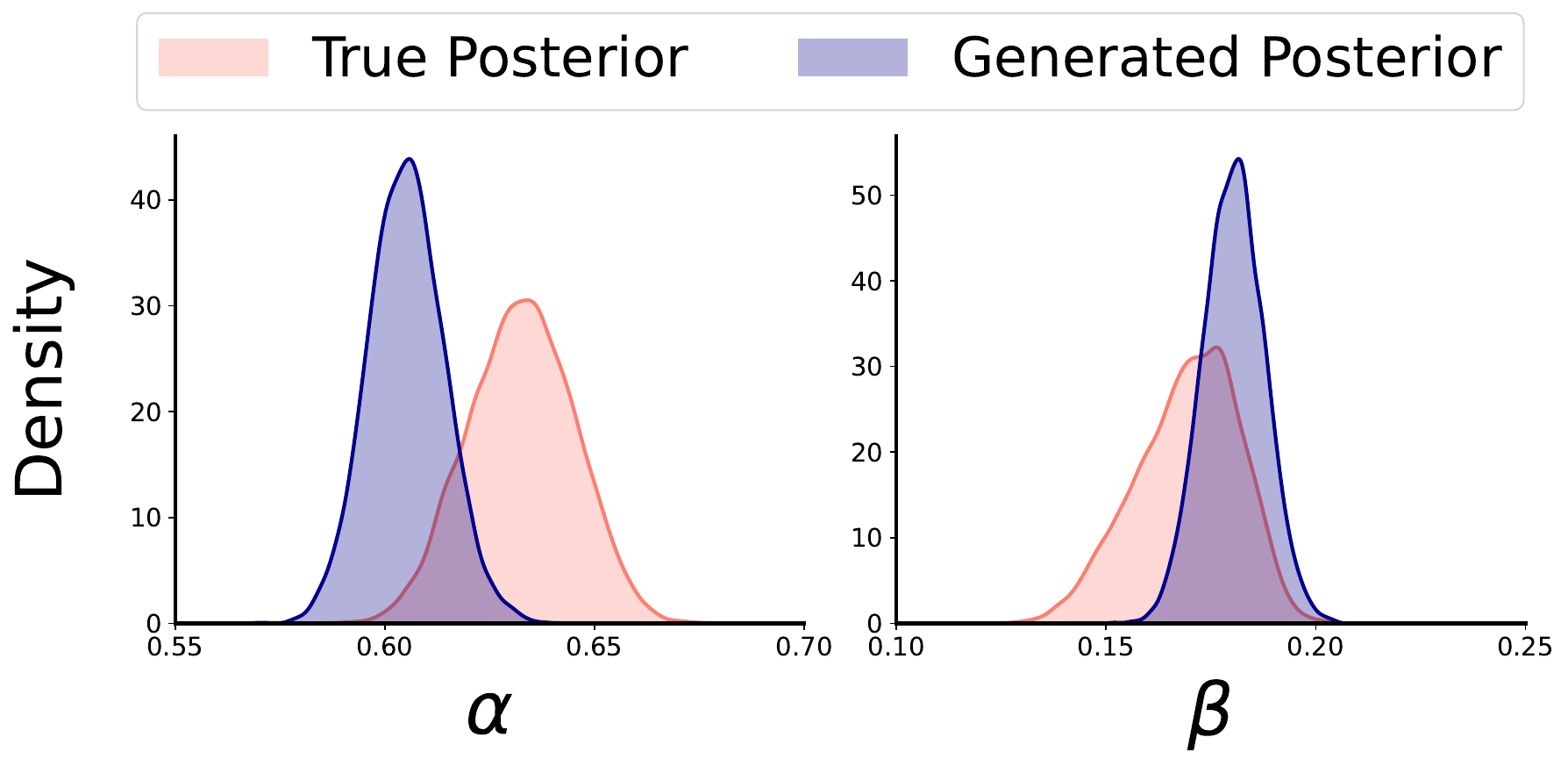}
        \caption{CP-VAE}
        \label{fig:sir_v1}
    \end{subfigure}
    \vfill
    \begin{subfigure}[t]{0.4\textwidth}
        \centering
        \includegraphics[width=\textwidth]{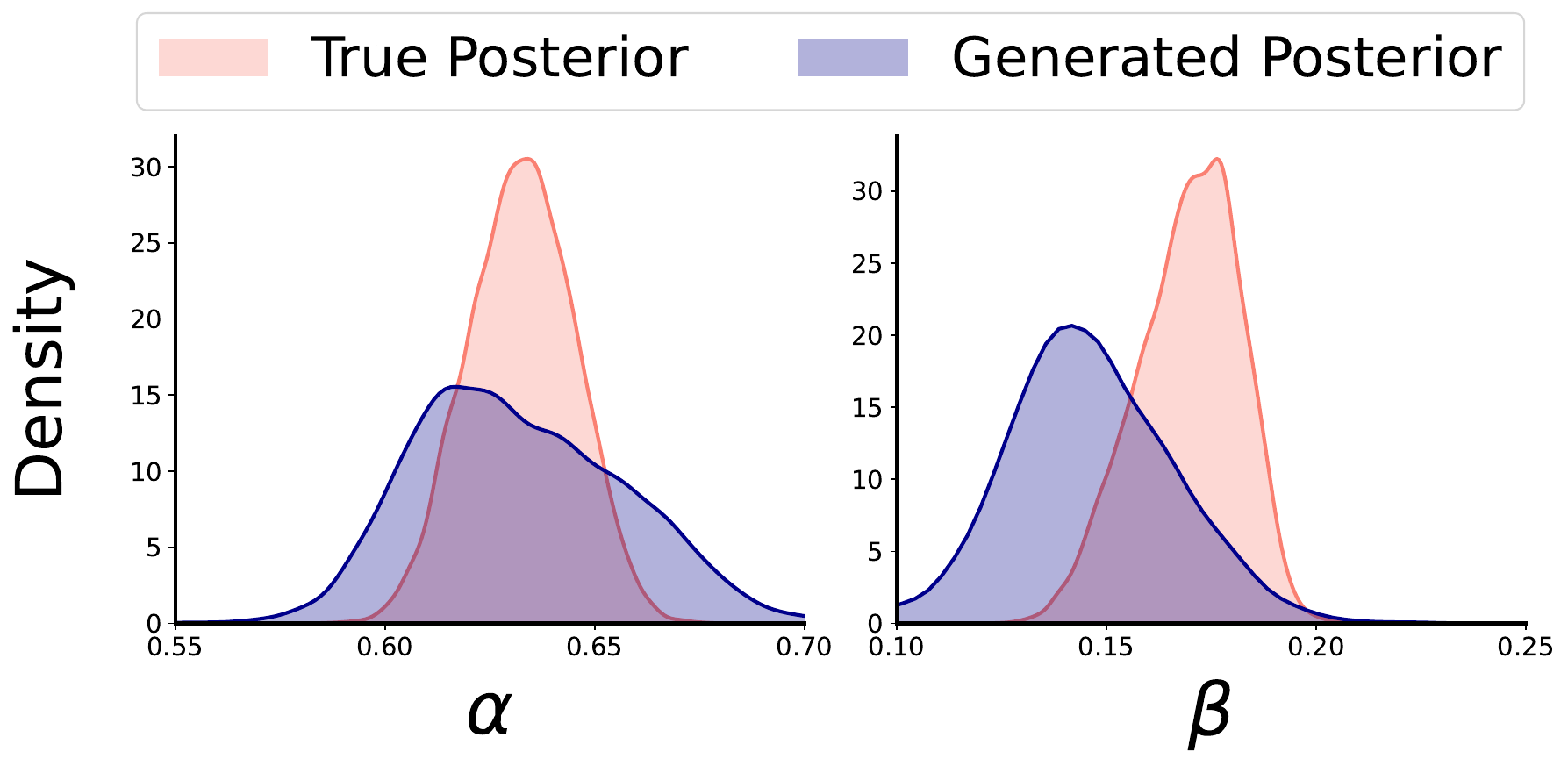}
        \caption{UP-VAE}
        \label{fig:sir_v2}
    \end{subfigure}
    \caption{SIR: Estimated posterior (simulation budget: 30,000). (a) CP-VAE and (b) UP-VAE.}
    \label{fig:sir}
\end{figure}

This section presents additional results for Bernoulli GLM Raw and SIR tasks. 
As Bernoulli GLM considers raw noisy data, CP-VAE's prior \( p(\mathbf{z}|\mathbf{y}) \) is affected, as \( \mathbf{y} \in \mathbb{R}^{100} \) contains significant noise, impacting training dynamics. 
In contrast, UP-VAE's prior \( p(\mathbf{z}) \) remains unaffected, leading to better performance (Figure~\ref{fig:glm_raw}). 
For SIR, both models sufficiently capture the parameters, though UP-VAE exhibits higher variance (Figure~\ref{fig:sir}).

\end{document}